\documentclass[10pt,twocolumn,letterpaper]{article}

\usepackage{iccv}              % To produce the CAMERA-READY version
\definecolor{iccvblue}{rgb}{0.21,0.49,0.74}
\usepackage[accsupp]{axessibility}
\usepackage[pagebackref,breaklinks,colorlinks,allcolors=iccvblue]{hyperref}
\usepackage{url}
\usepackage{graphicx}
\usepackage{subcaption}
\usepackage{amsmath}
\usepackage{algorithmicx}
\usepackage{algorithm}
\usepackage{algpseudocode}
\usepackage{mathrsfs}
\usepackage[graphicx]{realboxes}
\usepackage{multirow,booktabs}
\usepackage{wrapfig,lipsum,booktabs}
\usepackage{amsthm}
\usepackage[figuresright]{rotating}
\usepackage{bbm}
\usepackage{pgfplots}
\usepackage{enumitem}
\usepackage{tcolorbox}
\newtheorem{theorem}{Theorem}
\newtheorem{theorem1}{Theorem}
\newtheorem{theoremdef}{Theorem}
\newtheorem{definition}[theoremdef]{Definition}
\newtheorem{theoremre}{Theorem}
\newtheorem{remark}[theoremre]{Remark}
\newtheorem{definition*}{Problem}
\newtheorem{theoremlemm}{Theorem}
\newtheorem{lemma}[theoremlemm]{Lemma}
\newtheorem{theoremass}{Theorem}
\newtheorem{assumption}[theoremass]{Assumption}

\newtheorem{theorepro}{Theorem}
\newtheorem{proposition}[theorepro]{Proposition}
\algrenewcommand\algorithmicrequire{\textbf{Input:}}
\algrenewcommand\algorithmicensure{\textbf{Output:}}

\newcommand{\notprop}{\propto\kern-1\@ptsize pt \diagup}
\usepackage{color, colortbl}
\definecolor{LightCyan}{rgb}{0.88,1,1}

\def\paperID{278} % *** Enter the Paper ID here
\def\confName{ICCV}
\def\confYear{2025}

\title{On the Provable Importance of Gradients for\\ Language-assisted Image Clustering}

\author{Bo Peng, Jie Lu, Guangquan Zhang, Zhen Fang\thanks{Correspondence Author}\\
University of Technology Sydney\\
Sydney, Australia\\
}

\begin{document}
\maketitle
\begin{abstract}
This paper investigates the recently emerged problem of Language-assisted Image Clustering (LaIC), where textual semantics are leveraged to improve the discriminability of visual representations to facilitate image clustering. 
Due to the unavailability of true class names, one of core challenges of LaIC lies in how to filter positive nouns, i.e., those semantically close to the images of interest, from unlabeled wild corpus data.
Existing filtering strategies are predominantly based on the off-the-shelf feature space learned by CLIP; however, despite being intuitive, these strategies lack a rigorous theoretical foundation. 
To fill this gap, we propose a novel gradient-based framework, termed as GradNorm, which is theoretically guaranteed and shows strong empirical performance. In particular, we measure the positiveness of each noun based on the magnitude of gradients back-propagated from the cross-entropy between the predicted target distribution and the softmax output. Theoretically, we provide a rigorous error bound to quantify the separability of positive nouns by GradNorm and prove that GradNorm naturally subsumes existing filtering strategies as extremely special cases of itself. Empirically, extensive experiments show that GradNorm achieves the state-of-the-art clustering performance on various benchmarks. Code is publicly available at \href{https://github.com/60pen9/On-the-Provable-Importance-of-Gradients-for-Language-Assisted-Image-Clustering}{here}.
\end{abstract}
\section{Introduction}
\label{sec:intro}
As a fundamental problem in machine learning, image clustering~\cite{1} seeks to separate a set of unlabeled images into multiple groups such that images in the same group are semantically similar to each other. 
Due to its ability to reveal the inherent semantic structure underlying the data without requiring laborious and trivial data labeling work, clustering has been shown to benefit downstream tasks~\cite{2,3,4,c,d,e,f,g,h,i,j,k,l,m} in computer vision. 
Despite increasing attention, the vast majority of strategies~\cite{5,6,7,8,9,10,11} to image clustering reply on purely visual supervision signals and therefore inherit limitations especially when images of interest are visually similar to but semantically different from each other.

This paper delves into a new landscape for image clustering by departing from the classic single-model toward a multi-modal regime. 
In the visual domain, (deep) clustering methods usually learn discriminative representations from distributional priors~\cite{9,12,13}, pseudo-labels~\cite{14,15,16,17}, neighborhood consistency~\cite{18,19,20} and augmentation invariance~\cite{21,22,23}, which, however, can not be directly transferred into the vision-language regime due to the heterogeneous relation between visual and textual data.
While the advanced vision-language pre-training schemes, e.g., CLIP~\cite{24}, have emerged as promising alternatives for visual representation learning by mapping textual and visual inputs into a unified representation space, harnessing the power of texts to facilitate image clustering is still non-trivial due to the \textit{unavailability of class name priors}. 

To address this challenge, the mainstream solutions~\cite{26,27} are to select positive nouns, i.e., those who best describe images of interest, from unlabeled lexical databases in the wild\footnote{Generally, “in-the-wild” data are those that can be collected almost for free upon deploying machine learning models in the open world.}(e.g., WordNet~\cite{25}) for the textual pseudo-labeling of each image.
Despite recent empirical successes, the \textit{separability of positive nouns} remains theoretically underexplored, with no prior work providing a rigorous formalization or provable error bounds. Our work thus complements existing works by filling in the critical blank.
In this paper, we design a simple yet effective framework that provides a provable guarantee for Language-assisted Image Clustering (LaIC) from a novel perspective of gradient.

Methodologically, our proposed method GradNorm begins by learning a single-layer self-supervised classifier using CLIP features extracted from pseudo-labeled images. Leveraging the alignment between the CLIP image and text feature spaces~\cite{74}, we extend the learned classifier to handle text features as well. Subsequently, we employ CLIP features of unlabeled wild texts as input to compute the gradients of the classifier back-propagated based on the cross-entropy between the softmax output and the predicted target distribution. In this process, we consider unlabeled wild nouns as positive samples if the magnitude of the corresponding gradients falls below an adaptive threshold.

Theoretically, we justify GradNorm in Theorem~\ref{the1} and Section~\ref{sec4}. Our theoretical insights are twofold. First, we derive a rigorous upper bound on the error rate for separating positive nouns from unlabeled wild data. This upper bound is proportional to the optimal risk, which can approach zero in practice especially when the size of the pre-trained CLIP model is sufficiently large. Second, our analysis establishes a unified framework for existing filtering strategies~\cite{26,27} by demonstrating that, despite their apparent differences in motivation and methodology, they can be interpreted as degenerated cases of GradNorm.

Extensive experiments on multiple benchmarks demonstrate the empirical effectiveness of our proposed GradNorm method. For example, GradNorm achieves 60.6\% ACC and 81.2\% ACC on CIFAR-20 and ImageNet-Dog datasets, respectively, outperforming the latest TAC \cite{26} by 4.8\% and 6.1\%. Additionally, on three more challenging datasets (DTD, UCF-101, and ImageNet-1K), our method surpasses TAC \cite{26} by an average of 3.2\%, 1.7\%, and 2.4\% in terms of ACC, NMI, and ARI, respectively.

\section{Related Work}
\label{sec:formatting}
\subsection{Deep Image Clustering}
The popularity of deep image clustering can be attributed to the fact that distributional assumptions in classic clustering methods, e.g., compactness~\cite{28,guo1}, connectivity~\cite{29,b,guo2,guo3}, sparsity~\cite{30,a} and low rankness~\cite{31}, can not be necessarily conformed by high-dimensional structural RGB images. To exploit the powerful representative ability of deep neural networks in an unsupervised manner, the earliest attempts seeks self-supervision signals by considering image reconstruction~\cite{8,32,33,n}, probabilistic modeling~\cite{9,12,13} and mutual information maximization~\cite{34,35} as proxy tasks. Despite remarkable progresses, the learned representations may not be discriminative enough to capture the semantic similarity between images. More recently, the advance in self-supervised representation learning have led to major breakthroughs in deep image clustering. On the one hand, IDFD~\cite{36} proposes to perform both instance discrimination and feature de-correlation while MICE~\cite{37} propose a
unified latent mixture model based on contrastive learning to tackle the clustering task. On the other hand, CC~\cite{21} and its followers TCC~\cite{22} perform contrastive learning at both instance and cluster levels.
Different from above methods, ProPos~\cite{6} performs non-contrastive learning on the instance level and contrastive learning on the cluster level, which results in enjoying the strengths of both worlds.
\subsection{Vision-language Models}
Leveraging large-scale pre-trained vision-language models (VLMs) has emerged as a remarkably effective paradigm for multi-modal downstream tasks. Regarding the type of architectures, existing VLMs can be divided into two categories: 1) single-stream models like VisualBERT~\cite{38} and ViLT~\cite{39} feed the concatenated text and visual features into a single transformer-based encoder; 2) dual-stream models such as CLIP~\cite{24}, ALIGN~\cite{40}, and FILIP~\cite{41} use separate encoders for text and image and optimize with contrastive objectives to align semantically similar features in different modalities. In particular, CLIP enjoys popularity due to its simplicity and strong performance. CLIP-like models inspire numerous follow-up works~\cite{42,43,44} that aim to improve data efficiency and better adaptation to downstream tasks. This paper uses CLIP as the pre-trained model, but our method can be generally applicable to contrastive models that promote vision-language alignment.
\subsection{Language-assisted Image Clustering}
The core of LaIC lies in how to leverage texture semantics as the supervision signal to guide clustering in the visual domain. The seminar work called SIC~\cite{27} uses textual semantics to enhance image pseudo-labeling, followed by performing image clustering with consistency learning in both image space and semantic space. Note that, SIC essentially pulls image embeddings closer to embeddings in semantic space, while ignoring the improvement of text semantic embeddings. Differently, TAC~\cite{26} focuses on leveraging textual semantics to enhance the feature discriminability by either simply concentrating textual and visual features or its proposed cross-modal mutual distillation strategy. Despite their variety in the usage of texture semantics for image clustering, both SIC and TAC requires filter positive semantics from unlabeled wild textual data due to the lack of true class names. However, to the best of our knowledge, a formalized understanding regarding the separation of positive semantics is currently lacking for this field, which directly motivates our work.

\section{Proposed Framework: GradNorm}
% \textbf{Notations.} We use lowercase letters, lowercase boldface letters, and uppercase boldface letters to denote scalars, vectors, and matrices, respectively. Finally, we use $[N]$ to denote the index set $\left\{1,...,N\right\}$.. 
\subsection{Preliminary: Zero-shot Classification}
Let $\mathcal{X}$ and $\mathcal{T}$ be the visual and textual input space respectively, CLIP-based models adopt a simple dual-stream architecture with one text encoder $f_{\mathcal{T}}$ and one image encoder $f_{\mathcal{X}}$ to map inputs of two modalities into a uni-modal hyper-spherical feature space $\mathcal{Z}\triangleq\left \{\mathbf{z}\in\mathbb{R}^{d}|\left \|\mathbf{z}\right \|_2=1\right \}$. 
Considering an image classification task with the known classes $\left\{\mathbf{c}_1,\ldots,\mathbf{c}_K\right\}$, CLIP-based models make class prediction for any input image $\mathbf{x}\in\mathcal{X}$ by computing the following
\begin{equation}
\label{eq1}
\arg\max_{i=1,\ldots K}\frac{\exp\big[\tau f_{\mathcal{X}}(\mathbf{x})^\top f_{\mathcal{T}}\big(\Delta(\mathbf{c}_i)\big)\big]}{\sum_{j=1}^K\exp\big[\tau f_{\mathcal{X}}(\mathbf{x})^\top f_{\mathcal{T}}\big(\Delta(\mathbf{c}_j)\big)\big]},
\end{equation}
where $\tau>0$ is a temperature hyper-parameter, $\Delta(\mathbf{c}_i)\in\mathcal{T}$ with $\Delta(\cdot)$ as the prompt template for the input class name.
\subsection{Leveraging Unlabeled Textual Data in the Wild}
Despite remarkable effectiveness~\cite{24} and provable guarantees~\cite{45}, the zero-shot paradigm in Eq. (\ref{eq1}) suffers from the reliance on the prior knowledge of true class names, therefore inapplicable to the task of image clustering since we have access to only the number of ground-truth classes $K$.

In this paper, we address this challenge by leveraging unlabeled “in-the-wild” textual data which can be collected almost for free in the open world. However, it is important to note that wild textual data inevitably contains a mixture of \textit{positive}\footnote{By definition, \textit{positive} nouns are those semantically \textit{relevant/similar} to \textit{any} class in a dataset while \textit{negative} nouns are those semantically \textit{irrelevant/dissimilar} to \textit{all} the classes.} and \textit{negative semantics} regarding to the image dataset of interest. In view of this, we propose to use the Huber contamination model~\cite{46} to model the marginal distribution of the wild textual data as follows:
\begin{definition}[Wild Data Distribution]
\label{D1}
Let $\mathbb{P}_{\text{pos}}$ and $\mathbb{P}_{\text{neg}}$ be the distributions of positive and negative textual data defined over $\mathcal{T}$, respectively. According to the Huber contamination model~\cite{46}, we can model the unlabeled textual data distribution $\mathbb{P}_{\text{wild}}$ as follows:
\begin{equation}
\label{eq2}
\mathbb{P}_{\text{wild}}\triangleq\pi\cdot\mathbb{P}_{\text{pos}}+(1-\pi)\cdot\mathbb{P}_{\text{neg}},
\end{equation}
where $\pi\in(0,1]$ is typically unknown in practice.
\end{definition}
\begin{definition}[Empirical Wild Dataset]
\label{D2}
An empirical wild textual dataset $\mathcal{D}_{\mathcal{T}}$ is sampled independently and identically distributed (i.i.d.) from the wild data distribution $\mathbb{P}_{\text{wild}}$.
\end{definition}
\noindent Following prior works~\cite{26,27}, we simulate the wild dataset $\mathcal{D}_{\mathcal{T}}$ by resorting to the off-the-shelf WordNet~\cite{25}. In particular, let $\left\{\tilde{\mathbf{c}}_1,\ldots,\tilde{\mathbf{c}}_M\right\}$ be a pre-defined subset of nouns from WordNet, we can write $\mathcal{D}_{\mathcal{T}}=\left\{\tilde{\mathbf{t}}_i=\Delta(\tilde{\mathbf{c}}_i)\right\}_{i=1}^M$.
\begin{remark}
While wild textual data can be available in abundant without requiring human annotations, harnessing such data is non-trivial due to the lack of clear membership (either positive or negative) for textual data in $\mathcal{D}_{\mathcal{T}}$. Therefore, we aim to devise an automated strategy that estimates the membership for samples within the unlabeled textual data, therefore enabling the assistance of language for image clustering. In what follows, we describe these two stages in Section~\ref{sec3.3} and Section~\ref{sec3.4} respectively.
\end{remark}

\subsection{Filtering Candidate Positive Semantics}
\label{sec3.3}
\textbf{Overview.} To separate candidate positive semantics from the wild dataset $\mathcal{D}_{\mathcal{T}}$, we employ a level-set estimation based on the gradient information. The gradients are estimated from a classifier trained on the pseudo-labeled images. We describe the procedure formally below.
\subsubsection{Classifier Pre-training}
To realize the idea, let $\mathcal{D}_{\mathcal{X}}=\left\{\mathbf{x}_1,\ldots\mathbf{x}_N\right\}$ denotes the image dataset of interest, we begin with extracting features from CLIP-based models for images in the dataset $\mathcal{D}_{\mathcal{X}}$ to have $\mathbf{E}=(\mathbf{e}_1,\ldots,\mathbf{e}_N)\in\mathbb{R}^{d\times N}$ where $\mathbf{e}_i=f_{\mathcal{X}}(\mathbf{x}_i)\in\mathcal{Z}$ for each $i\in[N]:=\left\{1,...,N\right\}$. By performing a classical clustering algorithm, e.g., $k$-means, on the image feature matrix $\mathbf{E}$ to grouping given images into $C$ clusters, we can produce pseudo-label $y_i\in\mathcal{Y}\triangleq[C]$ for each image $\mathbf{x}_i\in\mathcal{D}_{\mathcal{X}}$ to learn a single-layer classifier $h(\cdot;\mathbf{W}):\mathcal{Z}\rightarrow\mathbb{R}^C$ parameterized by $\mathbf{W}=(\mathbf{w}_1,\ldots,\mathbf{w}_C)\in\mathbb{R}^{d\times C}$ with the following empirical risk minimization (ERM):
\begin{equation}
\label{eq3}
\mathbf{W}^\star=\arg\min_{\mathbf{W}\in\mathcal{W}}\frac{1}{N}\sum_{i=1}^N\ell\big(h(\mathbf{e}_i;\mathbf{W}),y_i\big),
\end{equation}
where $\mathcal{W}$ is the parameter space and $\ell\big(h(\mathbf{e}_i;\mathbf{W}),y_i\big)$ is the cross-entropy between the softmax output $h(\mathbf{x}_i;\mathbf{W})=\text{softmax}(\tau\cdot\mathbf{e}_i^\top\mathbf{W})$ and the pseudo-target distribution, i.e.,
\begin{equation}
\label{eq4}
\ell\big(h(\mathbf{e}_i;\mathbf{W}),y_i\big)\triangleq-\log\frac{\exp(\tau\mathbf{e}_i^\top\mathbf{w}_{y_i})}{{\sum}_{k\in[C]}\exp(\tau\mathbf{e}_i^\top\mathbf{w}_{k})}.
\end{equation}
\subsubsection{Membership Estimation via Gradient Norm}
Key to this step, we perform a scoring procedure to measure the positiveness of each text in the wild dataset $\mathcal{D}_{\mathcal{T}}$ to $\mathcal{D}_{\mathcal{X}}$, the image dataset of interest. To formulate
the score function $S$, we forward the feature of each text in the wild dataset $\mathcal{D}_{\mathcal{T}}$ into the learned classifier $h(\cdot;\mathbf{W}^\star)$ to calculate the gradients w.r.t. the classifier parameters $\mathbf{W}^\star$ by back-propagating the cross entropy between the softmax output and the predicted target distribution. In particular, let $\tilde{\mathbf{R}}=(\tilde{\mathbf{r}}_1,\ldots,\tilde{\mathbf{r}}_M)\in\mathbb{R}^{d\times M}$ as the textual feature matrix for the wild dataset $\mathcal{D}_{\mathcal{T}}$ where $\tilde{\mathbf{r}}_i=f_{\mathcal{T}}(\tilde{\mathbf{t}}_i)\in\mathcal{Z}$ for each $\tilde{\mathbf{t}}_i\in \mathcal{D}_{\mathcal{T}}$, we define the gradient matrix $\mathbf{G}$ as follows:
\begin{equation}
\label{eq5}
\mathbf{G}=\begin{bmatrix}
\partial \ell\big(h(\tilde{\mathbf{r}}_1;\mathbf{W}^\star),\tilde{y}_1\big)/\partial \mathbf{W}^\star\\
  \vdots  \\
  \partial \ell\big(h(\tilde{\mathbf{r}}_M;\mathbf{W}^\star),\tilde{y}_M\big)/\partial \mathbf{W}^\star
\end{bmatrix},
\end{equation}
where $\tilde{y}_i=\arg\min_{k\in[C]}\ell\big(h(\tilde{\mathbf{r}}_i;\mathbf{W}^\star),k\big)$. To assign the membership with $\tilde{\mathbf{t}}_i\in\mathcal{D}_{\mathcal{T}}$, we define the estimation score $S$ as follows
\footnote{We provide detailed deviation of the second step in the appendix}:
\begin{equation}
\label{eq6}
\begin{split}
S(\tilde{\mathbf{t}}_i)
&=\left \|\frac{\partial \ell\big(h(\tilde{\mathbf{r}}_i;\mathbf{W}^\star),\tilde{y}_i\big)}{\partial \mathbf{W}^\star} \right\|_F^2\\
&=\tau^2\cdot\left(\sum_{k\in[C]}\tilde{\pi}_{ik}^2+1-2\max_{j\in[C]}\tilde{\pi}_{ij}\right),
\end{split}
\end{equation}
where $\|\cdot\|_F$ denotes the Frobenius norm and 
\begin{equation}
\label{eq7}
\tilde{\pi}_{ij}=\frac{\exp(\tau\cdot\tilde{\mathbf{r}}_i^\top\mathbf{w}_j^\star)}{\sum_{k\in[C]}\exp(\tau\cdot\tilde{\mathbf{r}}_i^\top\mathbf{w}_k^\star)}, \quad \forall j\in[C].
\end{equation}
Finally, we can arrive at the (potentially noisy) set of candidate positive text semantics as follows:
% \begin{equation}
% \hat{\mathcal{D}}_{\mathcal{T}}^\text{pos}\triangleq\bigcup_{k=1}^{C}\left\{\tilde{t}_i\in\mathcal{D}_{\mathcal{T}}\text{: } S(\tilde{\mathbf{t}}_i)\geq T_k \text{ and } k=\arg\max_{j\in[L]}\tilde{p}_{ij} \right\}
% \end{equation}
\begin{equation}
\small
\label{eq8}
\hat{\mathcal{P}}_{\mathcal{T}}(k)\triangleq\left\{\tilde{\mathbf{t}}_i\in\mathcal{D}_{\mathcal{T}}\text{: } S(\tilde{\mathbf{t}}_i)\leq T_k \text{ and } \arg\max_{j\in[C]}\tilde{\pi}_{ij}=k \right\},
\end{equation}
where $T_k$ denotes the $\beta$-th smallest score of text semantics in the set $\left\{\tilde{\mathbf{t}}_i\in\mathcal{D}_{\mathcal{T}}\text{: } k=\arg\max_{j\in[C]}\tilde{\pi}_{ij} \right\}$.
In the following, our main theorem formally quantifies the separability of truly positive text semantics from the wild dataset $\mathcal{D}_\mathcal{T}$ by leveraging the filtering strategy in Eq. (\ref{eq8}).
% Section~\ref{} provides formal guarantees to rigorously justify that $\hat{\mathcal{D}}_{\mathcal{T}}^\text{pos}$ contains truly positive text semantics with a large probability.
\begin{theorem}\footnote{Due to space limitation, we defer detailed proofs in the appendix.} 
\label{the1}
Let us define the ground-truth set of truly positive semantics from the wild data as
$$\mathcal{P}_{\mathcal{T}}(k)=\left\{\tilde{\mathbf{t}}_i\in\mathcal{D}_\mathcal{T}:\tilde{\mathbf{t}}_i\sim\mathbb{P}_\text{pos}\text{ and }\arg\max_{j\in[C]}\tilde{\pi}_{ij}=k \right\}$$ and $\left|\mathcal{P}_{\mathcal{T}}(k)\right|=B_k$. Under mild assumptions (cf. Appendix. 3), i.e., the loss function $\ell$ is $\gamma$-smooth and the parameter space $\mathcal{W}$ is bounded, with the probability at least $0.97$, we have the following:
\begin{equation*}
\begin{split}
\text{Err}_\text{pos}(k)
&\triangleq\frac{\left|\left\{\tilde{\mathbf{t}}_i\in\mathcal{P}_{\mathcal{T}}(k):S(\tilde{\mathbf{t}}_i)>T_k\right\}\right|}{B_k}\\
&\leq\frac{2\gamma}{T_k}\left[\min_{\mathbf{W}\in\mathcal{W}}\Omega(\mathbf{W})+O(\sqrt{\frac{1}{B_k}})+O(\sqrt{\frac{1}{N}})\right],
\end{split}
\end{equation*}
where  $\Omega(\mathbf{W})=\mathbb{E}_{(\mathbf{z},y)\sim\mathbb{P}_{\mathcal{Z}\mathcal{Y}}}\ell\big(h(\mathbf{z};\mathbf{W}),y\big)$ is the expected risk and we use $O(\cdot)$ to hide universal constant factors.
\end{theorem}
\begin{remark}
Theorem~\ref{the1} states that, under mild assumptions, $\text{ERR}_\text{pos}(k)$ is upper-bounded. In particular, if the following two regulatory conditions hold: 1) the size of the image data $N$ and that of the wild textual data $B_k$ are sufficiently large; 2) the minimal expected risk $\min_{\mathbf{W}\in\mathcal{W}}\Omega(\mathbf{W})$ is sufficiently small, then the upper bound is also small. 
\end{remark}
\subsection{Clustering with Candidate Positive Semantics}
\label{sec3.4}
After extracting the candidate positive semantics set $\hat{\mathcal{D}}_{\mathcal{T}}^\text{pos}=\bigcup_{k=1}^{C}\hat{\mathcal{P}}_{\mathcal{T}}(k)$ from the wild textual dataset $\mathcal{D}_{\mathcal{T}}$, it is essential to design an effective collaboration mechanism between text semantics and image semantics for clustering. Given that the primary contribution of this paper is to reliably select positive semantics from unlabeled wild textual data, we adopt the same post-hoc collaboration strategy as \cite{26}.  

In particular, for each image $\mathbf{x}_i$, we build the corresponding text counterpart $\mathbf{v}_i$ by resorting to deep set representations~\cite{58,75}, i.e.,
\begin{equation}
\label{eq9}
\mathbf{v}_i=\sum_{\tilde{\mathbf{t}}_j\in\hat{\mathcal{D}}_{\mathcal{T}}^\text{pos}}\left(\frac{\exp(\mathbf{e}_i^\top\tilde{\mathbf{r}}_j/\kappa)}{\sum_{\tilde{\mathbf{t}}_k\in\hat{\mathcal{D}}_{\mathcal{T}}^\text{pos}}\exp(\mathbf{e}_i^\top\tilde{\mathbf{r}}_k/\kappa)}\cdot\tilde{\mathbf{r}}_j\right),
\end{equation}
where $\kappa>0$ is a temperature hyper-parameter. 

Finally, we compute the cluster assignment for the image dataset $\mathcal{D}_\mathcal{X}$ by applying $k$-means on the concatenated image-text features $\left\{[\mathbf{e}_i;\mathbf{v}_i]\in\mathbb{R}^{2d}\right\}_{i=1}^N$. For clarity, we summarize the details of GradNorm in Algorithm~\ref{alg1}.
\begin{algorithm}[t]
\caption{Pipeline of \textbf{GradNorm}} 
\label{alg1}
\begin{algorithmic}[1]
\Require{
Image features $\left\{\mathbf{e}_i\right\}_{i=1}^N$,
Text features $\left\{\tilde{r}_i\right\}_{i=1}^M$,
Randomly initialized parameters $\mathbf{W}$
}
\newline\Comment{\underline{\textit{Stage 1: Filtering Candidate Positive Semantics}}}
\State Apply $k$-means on image features $\left\{\mathbf{e}_i\right\}_{i=1}^N$ to obtain pseudo-labels $\left\{y_i\in[C]\right\}_{i=1}^N$
\State Obtain $\mathbf{W}^\star$ by performing ERM in Eq. (\ref{eq3})
% \State Compute $\tilde{y}_i=\underset{k\in[C]}{\arg\min} \ell\big(h(\tilde{\mathbf{r}}_i;\mathbf{W}^\star),k\big),~\forall i\in[M]$.
\State Compute $S(\tilde{\mathbf{t}}_i)$ in Eq. (\ref{eq6}) to obtain $\hat{\mathcal{P}}_{\mathcal{T}}(k)$ via Eq. (\ref{eq8})
\State Get candidate positive semantics $\hat{\mathcal{D}}_{\mathcal{T}}^\text{pos}=\bigcup_{k=1}^{C}\hat{\mathcal{P}}_{\mathcal{T}}(k)$.
\newline\Comment{\underline{\textit{Stage 2: Clustering with Candidate Positive Semantics}}}
\State Compute $\mathbf{v}_i$ for each $\mathbf{e}_i$ via Eq. (\ref{eq9})
\State Apply $k$-means on the concatenated image-text features $\left\{[\mathbf{e}_i;\mathbf{v}_i]\right\}_{i=1}^N$ to obtain final cluster assignment
\end{algorithmic} 
\end{algorithm}
% -------------------------------------------------------------------------
\section{Discussions}
\label{sec4}
In this section, we discuss the theoretical connection between our method between prior works~\cite{26,27} by showing that the latter can be explained as extremely special cases of the former though they indeed seem to be quite distinct regarding their proposed filtering strategies.
In particular, our theoretical analysis is motivated by SeCu~\cite{47} to consider training the classifier $h(\cdot;\mathbf{W})$ with the following objective:
\begin{equation}
\label{eq10}
\small
\hat{\ell}\big(h(\mathbf{e}_i;\mathbf{W}),y_i\big)\triangleq-\log\frac{\exp(\tau\mathbf{e}_i^\top\mathbf{w}_{y_i})}{\exp(\tau\mathbf{e}_i^\top\mathbf{w}_{y_i}) +\underset{k\neq y_i}{\sum}\exp(\tau\mathbf{e}_i^\top\hat{\mathbf{w}}_{k})},
\end{equation}
where $\hat{\mathbf{w}}=sg(\mathbf{w})$ with $sg(\cdot)$ as the stop-gradient operator.
\begin{remark}
Clearly, $\hat{\ell}$ in Eq. (\ref{eq10}) differs from the standard cross entropy in Eq. (\ref{eq4}) in that each weight vector $\mathbf{w}_k$ is only updated by image features whose pseudo label $y_i=k$. While it has been shown in SeCu \cite{47} that $\hat{\ell}$ in Eq. (\ref{eq10}) can be more stable than the standard cross entropy when the size of training batch is so small that the weight vector $\mathbf{w}_k$ is only updated by image features whose pseudo label $y_i\neq k$, we note that, since the memory complexity of the classifier $h(\cdot;\mathbf{W})$ is only $O(d\cdot C)$, training with large batches (e.g., $2048$) can be applicable in this paper.
\end{remark}
\begin{theorem}[\cite{47}]
\label{the2}
Let $\mathbf{W}^\star=(\mathbf{w}_1^\star,\ldots,\mathbf{w}_C^\star)$ be the empirical risk minimizer of the loss function in Eq. (\ref{eq4}) over the dataset $\left\{(\mathbf{e}_i,y_i)\right\}_{i=1}^N$. If we fix $\left\|\mathbf{w}_k\right\|_2=1$ for any $k\in[C]$, we then arrive at the closed form of $\mathbf{W}^\star$ given by:
\begin{equation}
\label{eq11}
\begin{split}
\mathbf{w}_j^\star&=\Lambda\left(\frac{\sum_{i:y_i=j}(1-\pi_{ij})\mathbf{e}_i}{\sum_{i:y_i=j}(1-\pi_{ij})}\right),
\end{split}
\end{equation}
where the operator $\Lambda(\cdot)$ denotes the $L_2$-normalizer and
\begin{equation*}
\pi_{ij}=\frac{\exp(\tau\mathbf{e}_i^\top\mathbf{w}^\star_{j})}{\exp(\tau\mathbf{e}_i^\top\mathbf{w}^\star_{j}) +\underset{k\neq y_i}{\sum}\exp(\tau\mathbf{e}_i^\top\hat{\mathbf{w}}^\star_{k})}.
\end{equation*}
\end{theorem}
\subsection{Connection to TAC ~\cite{26}}
In the extremely special case where $\tau\rightarrow0$, we have $\pi_{ij}\rightarrow1/C$ to approximate $\mathbf{w}_j^\star$ in Eq. (\ref{eq11}) as the center of image features that belongs to $k$-th cluster:
\begin{equation}
\label{eq12}
\begin{split}
\mathbf{w}_j^\star\rightarrow\Lambda\left(\frac{\sum_{i:y_i=j}\mathbf{e}_i}{\sum_{i}\mathbb{I}(y_i=j)}\right)~~\text{as}~~\tau\rightarrow0,
\end{split}
\end{equation}
where $\mathbb{I}(\cdot)$ is the indicator function. In this way, we can arrive at the same maximum softmax probability (MSP)-based filtering score used in TAC~\cite{26} as a special case of our proposed score in Eq. (\ref{eq6}), i.e.,  
\begin{equation}
\label{eq13}
\small
\begin{split}
\hat{S}(\tilde{\mathbf{t}}_i)
&=\left \|\frac{\partial \hat{\ell}\big(h(\tilde{\mathbf{r}}_i;\mathbf{W}^\star),\tilde{y}_i\big)}{\partial \mathbf{W}^\star} \right\|_F^2\\
&=\left \|\frac{\partial \ell\big(h(\tilde{\mathbf{r}}_i;\mathbf{W}^\star),\tilde{y}_i\big)}{\partial \mathbf{w}_{\tilde{y}_i}^\star} \right\|_2^2
\propto\left(1-\max_{j\in[C]}\tilde{\pi}_{ij}\right)^2,
\end{split}
\end{equation}
so that $\hat{S}(\tilde{\mathbf{t}}_i)\leq T_k\Leftrightarrow\underset{j\in[C]}{\max}\tilde{\pi}_{ij}\geq T_k'$ as $\max_{j\in[C]}\tilde{\pi}_{ij}\leq1$.
\begin{remark}
It is important to note that the effectiveness of MSP-based score function $\hat{S}$ in Eq. (\ref{eq13}) can be challenged by the notorious overconfidence phenomenon~\cite{48} where neural networks tend to produce overconfident predictions, i.e., abnormally high softmax confidences, even when the inputs are far away from the training data.
\end{remark}
\subsection{Connection to SIC~\cite{27}}
\begin{assumption}[Self-normalization~\cite{37,49}]
\label{A5}
An unnormalized classifier $h(\cdot,\mathbf{W})$ is self-normalized, i.e., for any possible input $z\in\mathcal{Z}$,
$\sum_{k\in[C]}\exp(\mathbf{z}^\top\mathbf{w}_k/\tau)=\text{const}$, so that
\begin{equation*}\tilde{\pi}_{ij}=\frac{\exp(\tau\tilde{\mathbf{r}}_i^\top\mathbf{w}_j^\star)}{\sum_{k\in[C]}\exp(\tau\tilde{\mathbf{r}}_i^\top\mathbf{w}_k^\star)}\propto\exp(\tau\tilde{\mathbf{r}}_i^\top\mathbf{w}_j^\star), \forall j\in[C].
\end{equation*}
\end{assumption}
\noindent In the extremely special case where $\tau\rightarrow0$, if Assumption~\ref{A5} holds for the classifier $h(\cdot,\mathbf{W}^\star)$ given by Eq. (\ref{eq12}), we have:
\begin{equation}
\label{eq14}
\arg\max_{j\in[C]}\tilde{\pi}_{ij}=\arg\max_{j\in[C]}\tilde{\mathbf{r}}_i^\top\mathbf{w}_j^\star.
\end{equation}
Combining Eq. (\ref{eq14}) and Eq. (\ref{eq13}), we can arrive at the same cosine similarity-based scoring function used in SIC~\cite{27} as a special case of our proposed score in Eq. (\ref{eq6}), i.e., 
\begin{equation}
\label{eq16}
\hat{S}(\tilde{\mathbf{t}}_i)\leq T_k\Leftrightarrow\underset{j\in[C]}{\max}\tilde{\pi}_{ij}\geq T_k'\Leftrightarrow\max_{j\in[C]}\tilde{\mathbf{r}}_i^\top\mathbf{w}_j^\star\geq T_k''.
\end{equation}

\begin{table*}[t]
\begin{center}
\caption{Clustering performance (\%) on five widely used image clustering datasets. The best results are highlighted in bold.}
\resizebox{\textwidth}{!}{%
\begin{tabular}{l|ccc|ccc|ccc|ccc|ccc}
\toprule
\multicolumn{1}{l|}{\textbf{Dataset}} 
                 & \multicolumn{3}{c|}{\textbf{STL-10}}   
                 & \multicolumn{3}{c|}{\textbf{CIFAR-10}} 
                 & \multicolumn{3}{c|}{\textbf{CIFAR-20}}                     & \multicolumn{3}{c|}{\textbf{ImageNet-10}}                  & \multicolumn{3}{c}{\textbf{ImageNet-Dogs}}                \\
\midrule
\textbf{Metrics} & \textbf{NMI} & \textbf{ACC} & \textbf{ARI}             
                 & \textbf{NMI} & \textbf{ACC} & \textbf{ARI}             
                 & \textbf{NMI} & \textbf{ACC} & \textbf{ARI}             
                 & \textbf{NMI} & \textbf{ACC} & \textbf{ARI}             
                 & \textbf{NMI} & \textbf{ACC} & \textbf{ARI}             
                 \\
\midrule
\rowcolor{gray!40} CLIP (zero-shot) & 93.9 & 97.1 & 93.7 & 80.7 & 90.0 & 79.3 & 55.3 & 58.3 & 39.8 & 95.8 & 97.6 & 94.9 & 73.5 & 72.8 & 58.2\\
\midrule
JULE (CVPR16)~\cite{59} & 18.2 & 27.7 & 16.4 & 19.2 & 27.2 & 13.8 & 10.3 & 13.7 & 3.3 & 17.5 & 30.0 & 13.8 & 5.4 & 13.8 & 2.8\\
DEC (ICML16)~\cite{8} & 27.6 & 35.9 & 18.6 & 25.7 & 30.1 & 16.1 & 13.6 & 18.5 & 5.0 & 28.2 & 38.1 & 20.3 & 12.2 & 19.5 & 7.9\\
DAC (ICCV17)~\cite{15} & 36.6 & 47.0 & 25.7 & 39.6 & 52.2 & 30.6 & 18.5 & 23.8 & 8.8 & 39.4 & 52.7 & 30.2 & 21.9 & 27.5 & 11.1\\
DCCM (ICCV19)~\cite{60} & 37.6 & 48.2 & 26.2 & 49.6 & 62.3 & 40.8 & 28.5 & 32.7 & 17.3 & 60.8 & 71.0 & 55.5 & 32.1 & 38.3 & 18.2\\
IIC (ICCV19)~\cite{35} & 49.6 & 59.6 & 39.7 & 51.3 & 61.7 & 41.1 & 22.5 & 25.7 & 11.7 & — & — & — & — & — & — \\
PICA (CVPR20)~\cite{16} & 61.1 & 71.3 & 53.1 & 59.1 & 69.6 & 51.2 & 31.0 & 33.7 & 17.1 & 80.2 & 87.0 & 76.1 & 35.2 & 35.3 & 20.1\\
CC (AAAI21)~\cite{21} & 76.4 & 85.0 & 72.6 & 70.5 & 79.0 & 63.7 & 43.1 & 42.9 & 26.6 & 85.9 & 89.3 & 82.2 & 44.5 & 42.9 & 27.4\\
IDFD (ICLR20)~\cite{36} & 64.3 & 75.6 & 57.5 & 71.1 & 81.5 & 66.3 & 42.6 & 42.5 & 26.4 & 89.8 & 95.4 & 90.1 & 54.6 & 59.1 & 41.3\\
SCAN (ECCV20)~\cite{18} & 69.8 & 80.9 & 64.6 & 79.7 & 88.3 & 77.2 & 48.6 & 50.7 & 33.3 & —    & —    & —    & 61.2 & 59.3 & 45.7\\
MiCE (ICLR20)~\cite{37} & 63.5 & 75.2 & 57.5 & 73.7 & 83.5 & 69.8 & 43.6 & 44.0 & 28.0 & —    & —    & —    & 42.3 & 43.9 & 28.6\\
GCC (ICCV21)~\cite{20} & 68.4 & 78.8 & 63.1 & 76.4 & 85.6 & 72.8 & 47.2 & 47.2 & 30.5 & 84.2 & 90.1 & 82.2 & 49.0 & 52.6 & 36.2\\
NNM (CVPR21)~\cite{19} & 66.3 & 76.8 & 59.6 & 73.7 & 83.7 & 69.4 & 48.0 & 45.9 & 30.2 & —    & —    & —    & 60.4 & 58.6 & 44.9\\
CRLC (ICCV21)~\cite{23} & 72.9 & 81.8 & 62.8 & 67.9 & 79.9 & 63.4 & 41.6 & 42.5 & 26.3 & 83.1 & 85.4 & 75.9 & 48.4 & 46.1 & 59.7\\
TCC (NeurIPS21)~\cite{58} & 73.2 & 81.4 & 68.9 & 79.0 & 90.6 & 73.3 & 47.9 & 49.1 & 31.2 & 84.8 & 89.7 & 82.5 & 55.4 & 59.5 & 41.7\\
TCL (IJCV22)~\cite{22} & 79.9 & 86.8 & 75.7 & 81.9 & 88.7 & 78.0 & 52.9 & 53.1 & 35.7 & 87.5 & 89.5 & 83.7 & 62.3 & 64.4 & 51.6\\
SPICE (TIP22)~\cite{62} & 81.7 & 90.8 & 81.2 & 73.4 & 83.8 & 70.5 & 44.8 & 46.8 & 29.4 & 82.8 & 92.1 & 83.6 & 57.2 & 64.6 & 47.9\\
SeCu (ICCV23)~\cite{47} & 70.7 & 81.4 & 65.7 & 79.9 & 88.5 & 78.2 & 51.6 & 51.6 & 36.0 & — & — & — & — & — & —\\
DivClust (CVPR23)~\cite{61} & — & — & — & 71.0 & 81.5 & 67.5 & 44.0 & 43.7 & 28.3 & 85.0 & 90.0 & 81.9 & 51.6 & 52.9 & 37.6\\
% CoNR (NeurIPS23) & 84.6 & 92.2 & 83.8 & 87.1 & 93.3 & 86.5 & 60.3 & 59.0 & 44.8 & 89.8 & 95.8 & 90.9 & 74.2 & 80.2 & 67.6\\
RPSC (AAAI24)~\cite{63} & 83.8 & 92.0 & 83.4 & 75.4 & 85.7 & 73.1 & 47.6 & 51.8 & 34.1 & 83.0 & 92.7 & 85.8 & 55.2 & 64.0 & 46.5\\
% IDC+TCL (NeurIPS24) & 85.3 & 92.7 & 84.6 & 84.4 & 92.7 & 84.8 & 58.1 & 69.4 & 48.7 & 93.2 & 97.2 & 93.9 & 69.1 & 78.8 & 63.6\\
\midrule
CLIP ($k$-means) & 91.7 & 94.3 & 89.1 & 70.3 & 74.2 & 61.6 & 49.9 & 45.5 & 28.3 & 96.9 & 98.2 & 96.1 & 39.8 & 38.1 & 20.1\\
SIC (AAAI23)~\cite{27} & 95.3 & 98.1 & 95.9 & {\bf 84.7} & {\bf 92.6} & {\bf 84.4} & 59.3 & 58.3 & 43.9 & 97.0 & 98.2 & 96.1 & 69.0 & 69.7 & 55.8\\
TAC (ICML24)~\cite{26} & 92.3 & 94.5 & 89.5 & 80.8 & 90.1 & 79.8 & 60.7 & 55.8 & 42.7 & 97.5 & 98.6 & 97.0 & 75.1 & 75.1 & 63.6\\
\rowcolor{LightCyan} GradNorm (ours)  &{\bf 95.6} & {\bf 98.3} & {\bf 96.2} & 82.6  & 91.1 & 81.5 & {\bf 61.3} & {\bf 60.6} & {\bf 43.6} & {\bf 98.7} & {\bf 99.4} & {\bf 98.7} & {\bf 81.0} & {\bf 81.2} & {\bf 70.9}\\
\bottomrule
\end{tabular}
}
\label{tab1}
\end{center}
\end{table*}

\begin{table*}[t]
\begin{center}
\caption{Clustering performance (\%) on three challenging image clustering datasets. The best results are highlighted in bold.}
\resizebox{0.8\textwidth}{!}{%
\begin{tabular}{l|ccc|ccc|ccc|ccc}
\toprule
\multicolumn{1}{l|}{\textbf{Dataset}} 
                 & \multicolumn{3}{c|}{\textbf{DTD}}   
                 & \multicolumn{3}{c|}{\textbf{UCF-101}} 
                 & \multicolumn{3}{c|}{\textbf{ImageNet-1K}}   
                 & \multicolumn{3}{c}{\textbf{Average}}  
                 \\
\midrule
\textbf{Metrics} & \textbf{NMI} & \textbf{ACC} & \textbf{ARI}             
                 & \textbf{NMI} & \textbf{ACC} & \textbf{ARI}             
                 & \textbf{NMI} & \textbf{ACC} & \textbf{ARI}     
                 & \textbf{NMI} & \textbf{ACC} & \textbf{ARI}
                 \\
\midrule
\rowcolor{gray!40} CLIP (zero-shot) & 56.5 & 43.1 & 26.9 & 79.9 & 63.4 & 50.2 & 81.0 & 63.6 & 45.4 & 72.5 & 56.7 & 40.8\\
\midrule
SCAN (ECCV20)~\cite{18} & 59.4 & 46.4 & 31.7 & 79.7 & 61.1 & 53.1 & 74.7 & 44.7 & 32.4 & 71.3 & 50.7 & 39.1\\
CLIP ($k$-means) & 57.3 & 42.6 & 27.4 & 79.5 & 58.2 & 47.6 & 72.3 & 38.9 & 27.1 & 69.7 & 46.6 & 34.0\\
SIC (AAAI23)~\cite{27} & 59.6 & 45.9 & 30.5 & 81.0 & 61.9 & 53.6 & 77.2 & 47.0 & 34.3 & 72.6 & 51.6 & 39.5\\
TAC (ICML24)~\cite{26} & 60.1 & 45.9 & 29.0 & 81.6 & 61.3 & 52.4 & 77.8 & 48.9 & 36.4 & 73.2 & 52.0 & 39.3\\
\rowcolor{LightCyan} GradNorm (ours)  &\textbf{63.1} & \textbf{50.9} & \textbf{34.2} & \textbf{82.5} & \textbf{62.7} & \textbf{53.2} & \textbf{79.2} & \textbf{52.6}  & \textbf{39.1} & \textbf{74.9} & \textbf{55.4} & \textbf{41.7}\\
\bottomrule
\end{tabular}
}
\label{tab2}
\end{center}
\end{table*}
\section{Experiments}
\subsection{Experimental Setups}
\subsubsection{Datasets}
We evaluate the effectiveness of GradNorm by conducting experiments on 1) five widely-used datasets: STL-10~\cite{50}, CIFAR-10~\cite{51}, CIFAR-20~\cite{51}, ImageNet-10~\cite{15}, and ImageNet-Dogs~\cite{15}; 2) three more complex datasets with larger cluster numbers: DTD~\cite{52}, UCF-101~\cite{53}, and ImageNet-1K~\cite{54}. 
Following prior works~\cite{26,27}, we filter candidate positive semantics based on the train split of each image dataset, followed by evaluate the clustering performance on the test split of each image dataset. To keep the main content concise, We summarize the details of these datasets in the appendix.
\subsubsection{Implementation Details}
For a fair comparison with previous works~\cite{24,26,27}, we, unless
explicitly stated, adopt the
pre-trained CLIP model with ViT-B/32~\cite{55} and Transformer~\cite{56} as default image and text backbones, respectively. 
For nouns from WordNet~\cite{25}, we assemble them with prompts like
“A photo of [CLASS]” before feeding them into the Transformer. 
To find semantics of appropriate granularity given a $N$-sized image dataset, we, similar to TAC~\cite{26}, set $C=N/600$ for datasets with an average cluster size larger than 600 and $C=3K$ otherwise.
We fix $\tau=1/0.08$, $\kappa=0.006$ and $\beta=5$ for all datasets.
In most cases, we train the classifier $h$ by the Adam~\cite{57} optimizer for 30 epochs with learning rate as $1e-3$ and batch size as 2048. The only exception is that on UCF-101 and ImageNet-1K, where, the classifier $h$ is trained for 100 epochs with batch size as 8192. 
All experiments are conducted on a single Nvidia A100 GPU.
\subsubsection{Evaluation Metrics}
We measure clustering performance by three metrics, including Accuracy (ACC), Normalized Mutual Information (NMI) and Adjusted Rand Index (ARI). 
The higher values of these metric indicates a better clustering performance.
% \subsubsection{Baseline Methods}
% We compare GradNorm with JULE~\cite{59}, DEC~\cite{8}, DAC~\cite{15}, DCCM~\cite{60}, IIC~\cite{35}, PICA~\cite{16}, CC~\cite{21}, IDFD~\cite{36}, SCAN~\cite{18}, MiCE~\cite{37}, GCC~\cite{20}, NNM~\cite{19}, CRLC~\cite{23}, SeCu~\cite{47}, DivClust~\cite{61},  RPSC~\cite{63}, TCL~\cite{62}, TCC~\cite{58}, SPICE~\cite{62}, SIC~\cite{27} and TAC~\cite{26}.

\begin{figure*}
\centering
% \captionsetup[subfigure]{labelformat=simple}
\begin{subfigure}{0.33\textwidth}
\includegraphics[width=\textwidth]{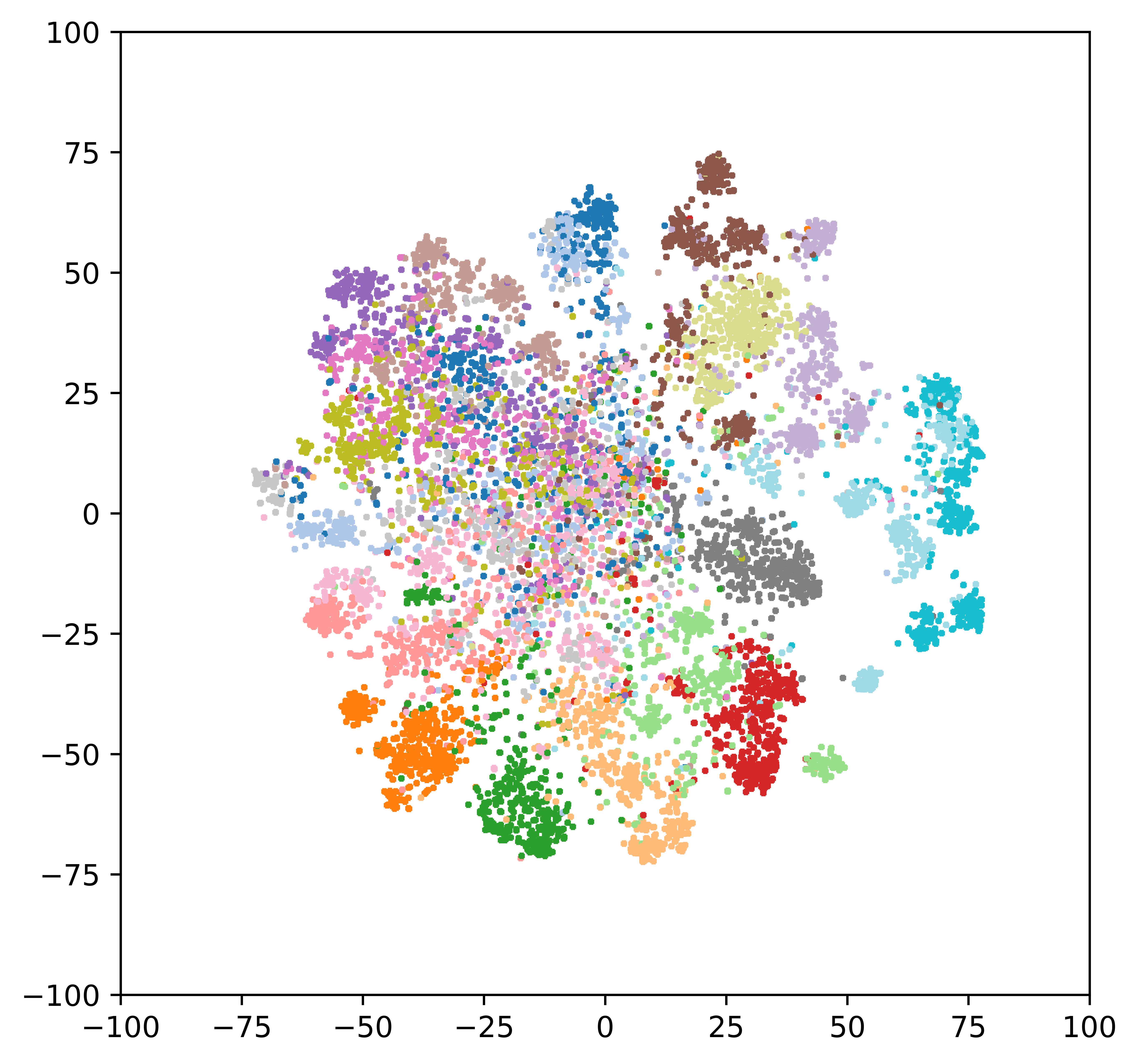}
\caption{CLIP Image features}
\label{fig1a}
\end{subfigure}
\hfill
\begin{subfigure}{0.33\textwidth}
    \includegraphics[width=\textwidth]{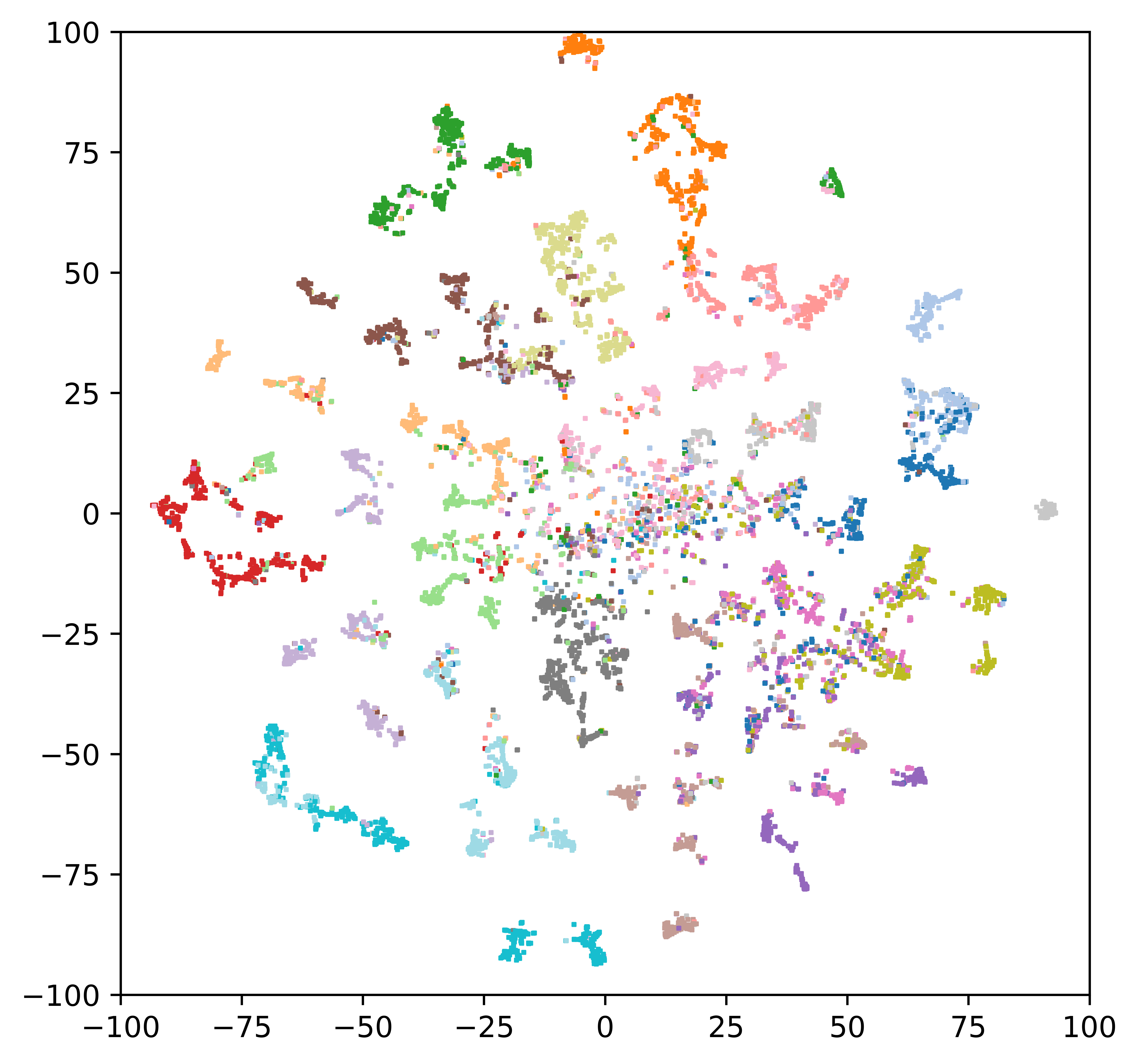}
    \caption{Text Counterpart}
    \label{fig1b}
\end{subfigure}
\hfill
\begin{subfigure}{0.33\textwidth}
    \includegraphics[width=\textwidth]{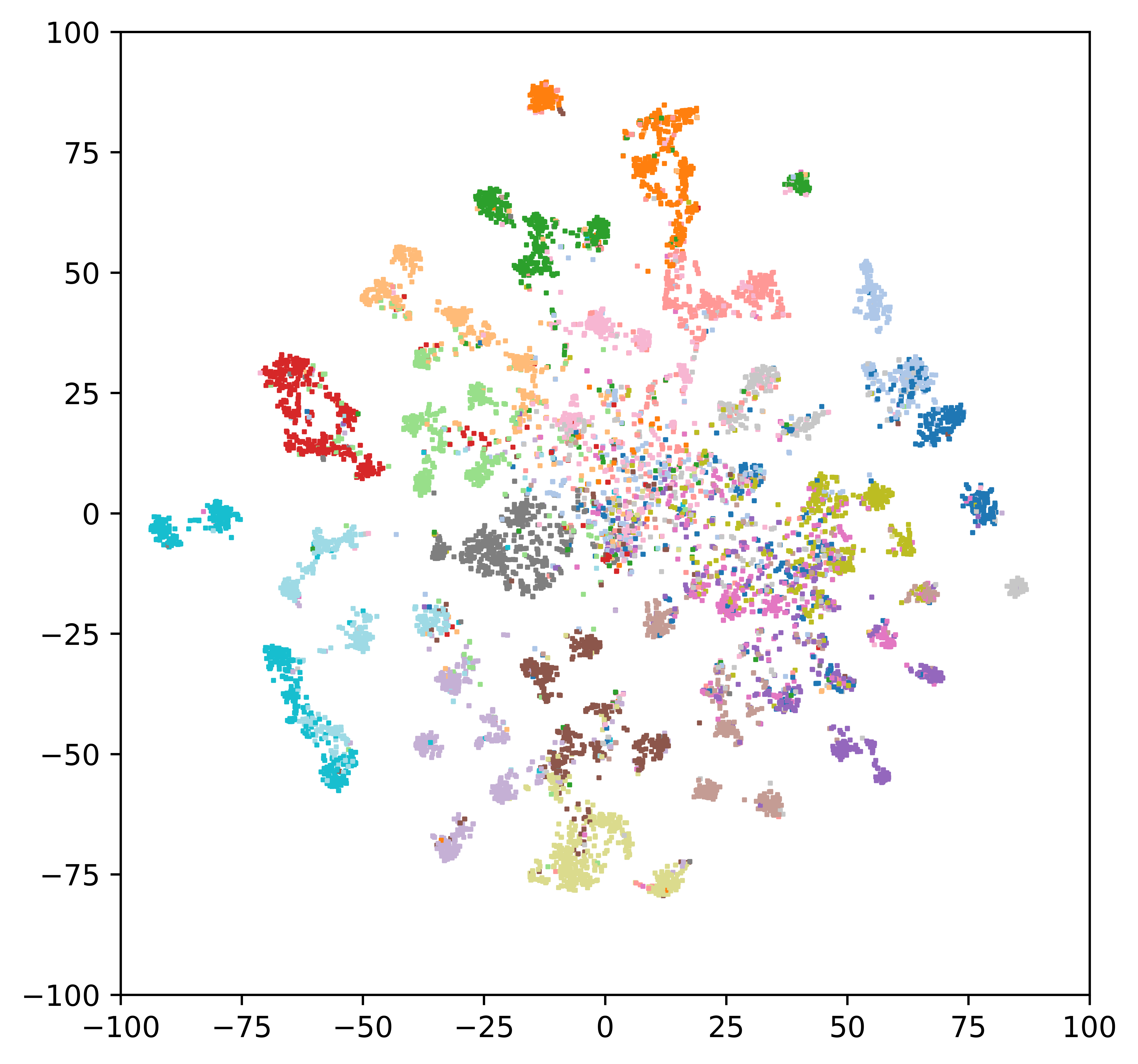}
    \caption{Concatenated Features}
    \label{fig1c}
\end{subfigure}
% \begin{subfigure}{0.24\textwidth}
%     \includegraphics[width=\textwidth]{ICCV2025-Author-Kit/sec/Figure_03.png}
%     \caption{}
%     \label{fig:third}
% \end{subfigure}
\caption{t-SNE Visualization of features extracted by different methods on the test split of CIFAR-20: (a) image embedding directly extracted from the pre-trained CLIP visual encoder, (b) text counterparts constructed
by the candidate semantics selected by Gradnorm, and (c) concatenation of images and text counterparts. Various colors indicate different ground-truth class assignment.}
\end{figure*}

% \begin{figure*}[h]
%   \centering
% \includegraphics[width=\linewidth]{ICCV2025-Author-Kit/sec/resnet.png}
%   \caption{Clustering peformance on five image clustering datasets, where CLIP-B/16 is used as encoder.}
% \label{r1}
% \end{figure*}

% \begin{figure*}[h]
%   \centering
% \includegraphics[width=\linewidth]{ICCV2025-Author-Kit/sec/resnet2.png}
%   \caption{Clustering peformance on five image clustering datasets, where CLIP-L/14 is used as encoder.}
% \label{r2}
% \end{figure*}

\begin{figure}[t!]
    \centering
    % \captionsetup[subfigure]{labelformat=simple}
    \begin{subfigure}[t]{0.25\textwidth}
        \centering
        \includegraphics[width=\textwidth]{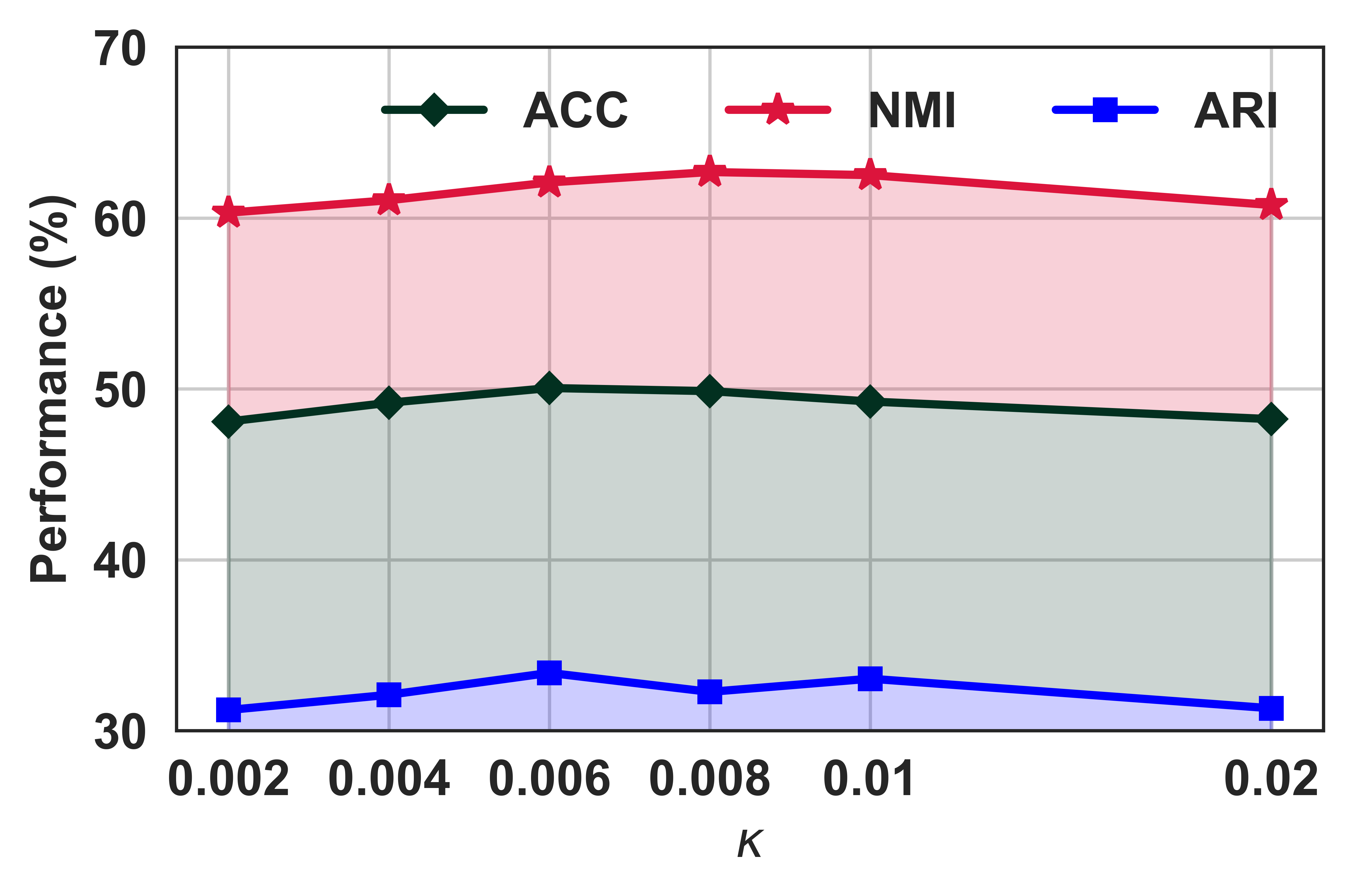}
        \caption{DTD}
        \label{fig0a}
    \end{subfigure}%
    \begin{subfigure}[t]{0.25\textwidth}
        \centering
        \includegraphics[width=\textwidth]{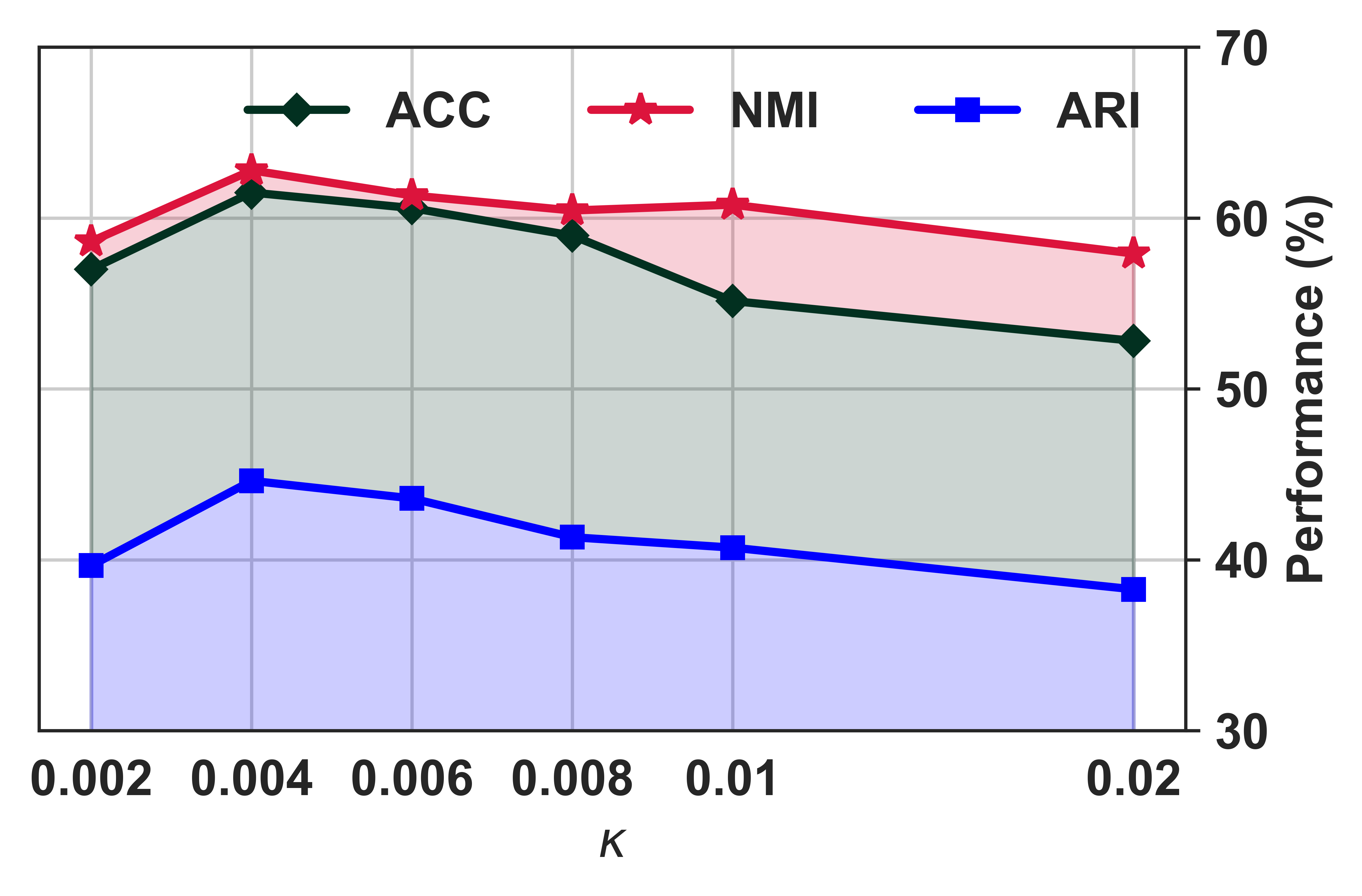}
        \caption{CIFAR-20}
        \label{fig0b}
    \end{subfigure}
    \caption{Analysis of clustering performance by varying the value of the temperature hyper-parameter $\kappa$ on (a) DTD and (b) CIFAR-20 datasets, respectively.}
\end{figure}

\begin{figure}[t!]
    \centering
    % \captionsetup[subfigure]{labelformat=simple}
    \begin{subfigure}[t]{0.25\textwidth}
        \centering
        \includegraphics[width=\textwidth]{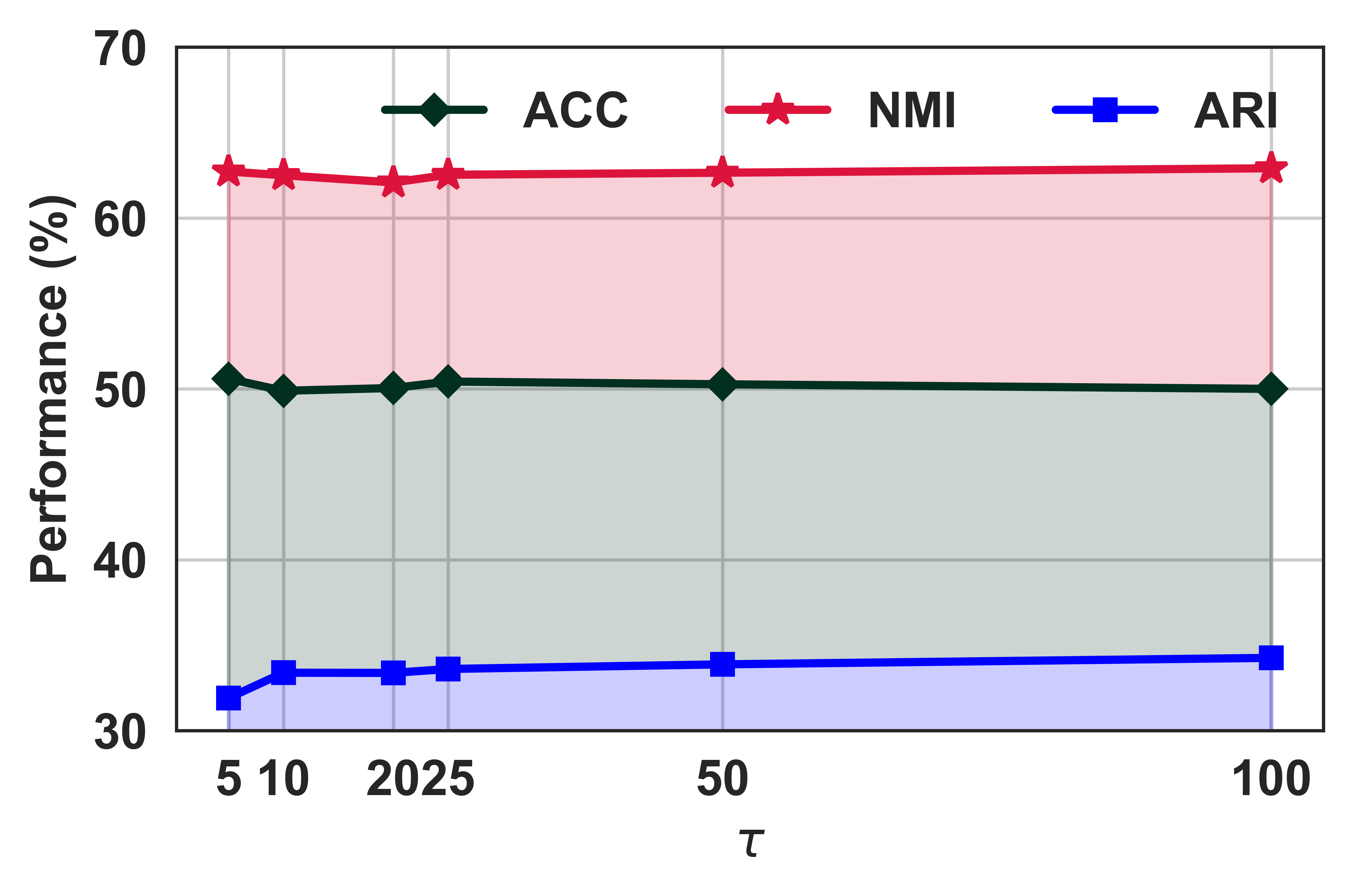}
        \caption{DTD}
        \label{fig1a}
    \end{subfigure}%
    \begin{subfigure}[t]{0.25\textwidth}
        \centering
        \includegraphics[width=\textwidth]{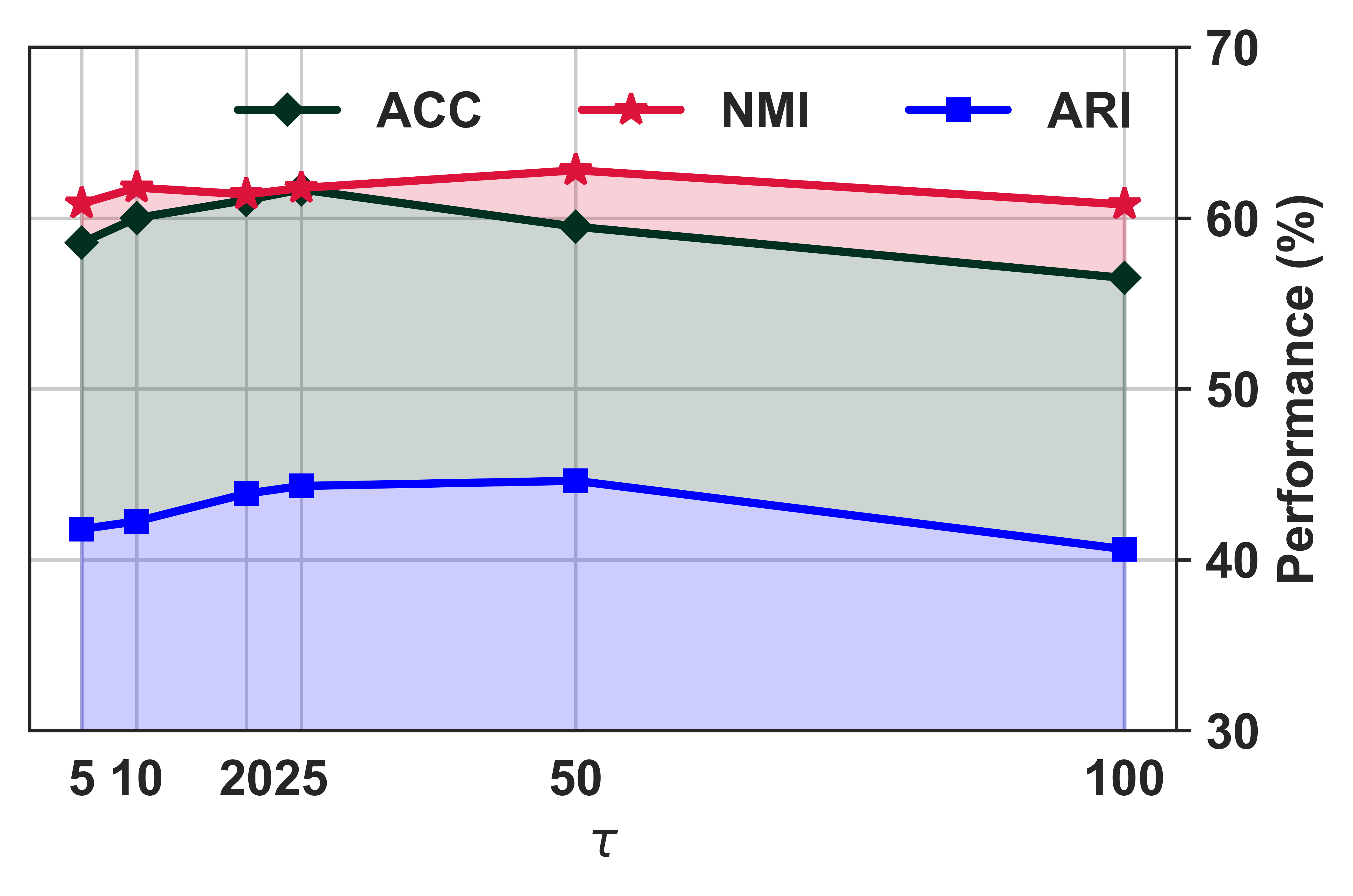}
        \caption{CIFAR-20}
        \label{fig1b}
    \end{subfigure}
    \caption{Analysis of clustering performance by varying the value of the temperature hyper-parameter $\tau$ on (a) DTD and (b) CIFAR-20 datasets, respectively.}
\end{figure}

\begin{figure}[t!]
    \centering
    % \captionsetup[subfigure]{labelformat=simple}
    \begin{subfigure}[t]{0.25\textwidth}
        \centering
        \includegraphics[width=\textwidth]{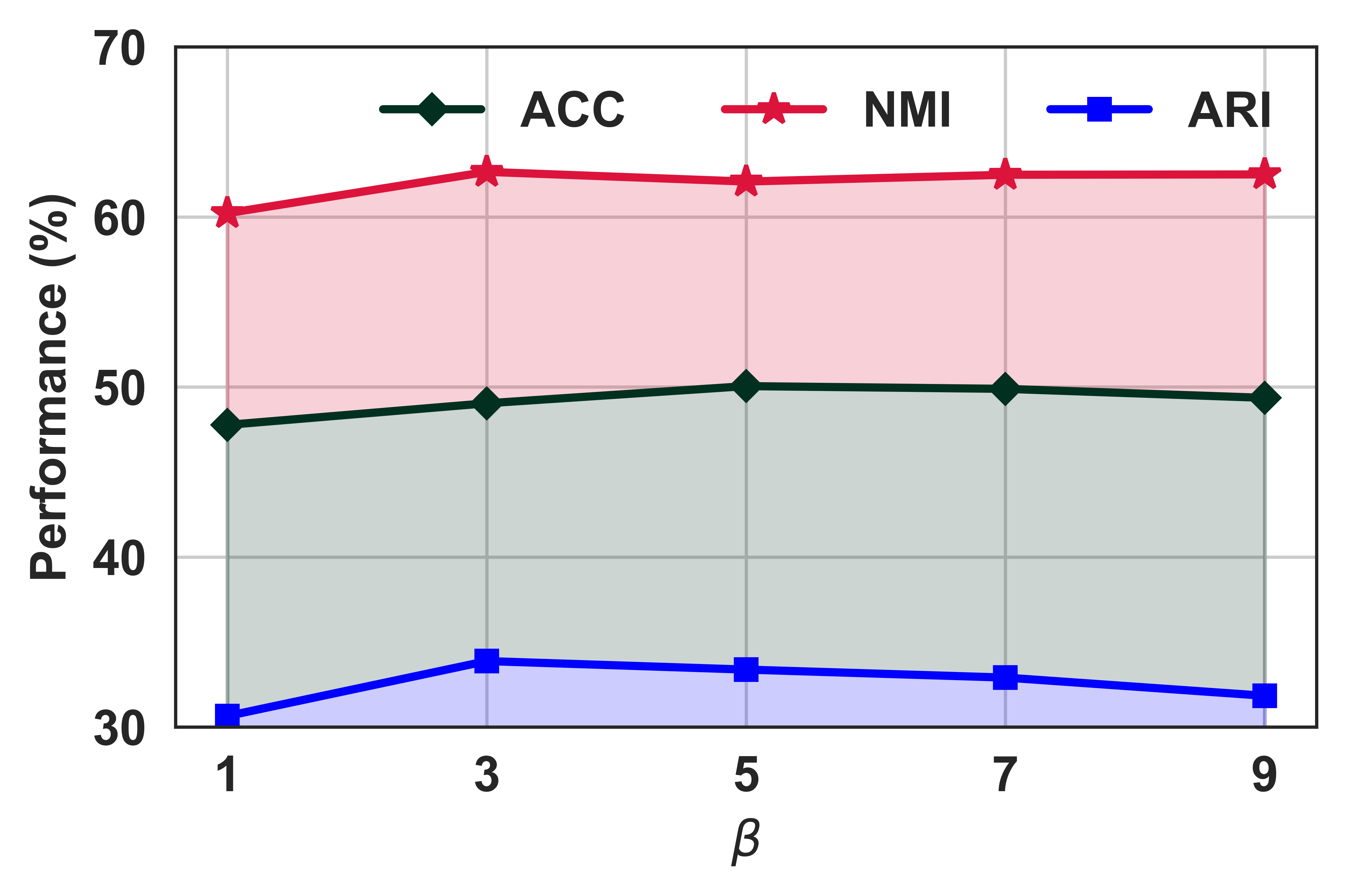}
        \caption{DTD}
        \label{fig2a}
    \end{subfigure}%
    \begin{subfigure}[t]{0.25\textwidth}
        \centering
        \includegraphics[width=\textwidth]{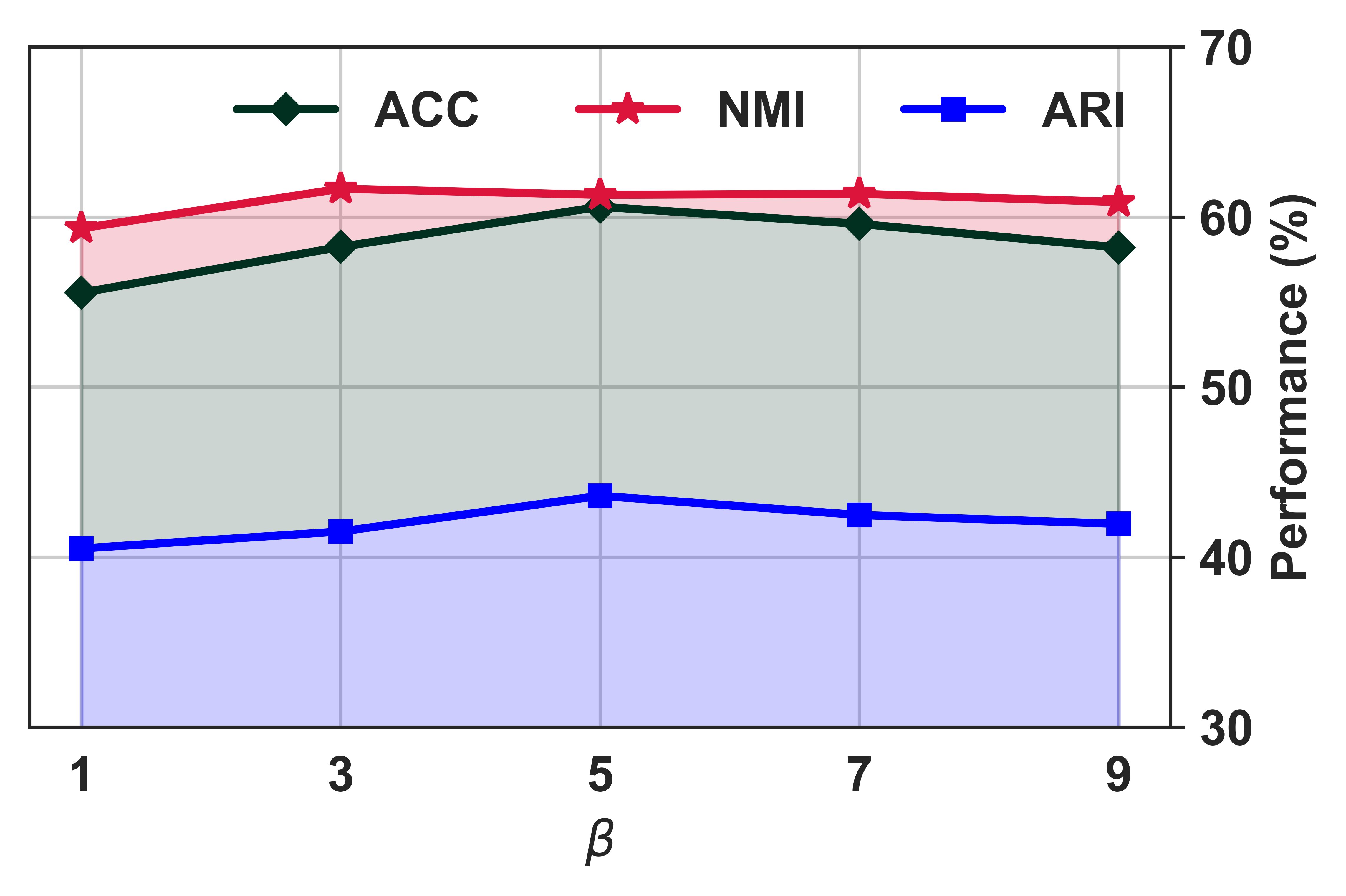}
        \caption{CIFAR-20}
        \label{fig2b}
    \end{subfigure}
    \caption{Analysis of clustering performance by varying $\beta$, the number of selected positive semantics, on (a) DTD and (b) CIFAR-20 datasets, respectively.}
\end{figure}

\begin{table*}[t]
\begin{center}
\caption{Clustering performance (\%) on five widely used image clustering datasets. The best results are highlighted in bold.}
\resizebox{0.8\textwidth}{!}{%
\begin{tabular}{l|l|ccc|ccc|ccc}
\toprule
\multicolumn{1}{l|}{\textbf{Visual}} 
                 & \multicolumn{1}{l|}{\textbf{Dataset}} 
                 & \multicolumn{3}{c|}{\textbf{CIFAR-10}} 
                 & \multicolumn{3}{c|}{\textbf{CIFAR-20}}             
                 & \multicolumn{3}{c}{\textbf{DTD}}                
                 \\
\cline{2-11}
\multicolumn{1}{l|}{\textbf{Backbone}} &\textbf{Metrics} 
                 & \textbf{NMI} & \textbf{ACC} & \textbf{ARI}             
                 & \textbf{NMI} & \textbf{ACC} & \textbf{ARI}             
                 & \textbf{NMI} & \textbf{ACC} & \textbf{ARI}             
                 \\
\midrule
\multirow{2}{*}{ViT-B/16} 
&TAC (ICML24) & 81.8 & 89.7 & 79.3 & 62.2 & 56.2 & 45.4  & 62.6 & 50.4 & 33.6\\
& GradNorm (ours)  & {\bf 83.6}  & {\bf 90.6} & {\bf 81.0} & {\bf 65.6} & {\bf 61.2} & {\bf 46.3} & {\bf 63.9} & {\bf 52.0} & {\bf 35.1}\\
\midrule
\multirow{2}{*}{ViT-L/14} 
&TAC (ICML24) & 89.1 & 93.9 & 86.7 & 66.1 & 57.8 & 45.3 & 65.5 & 51.6 & 34.0\\
& GradNorm (ours)  & {\bf 91.7} & {\bf 95.3} & {\bf 89.5} & {\bf 69.5} & {\bf 61.7} & {\bf 48.9} & {\bf 66.6} & {\bf 54.3} & {\bf 36.1}\\
\midrule
\multirow{2}{*}{ResNet-50} 
&TAC (ICML24) & 57.1 & 69.5 & 47.8 & 42.5 & 43.3 & 24.9 & 58.9 & 46.1 & 29.2\\
& GradNorm (ours)  & {\bf 60.6} & {\bf 72.2} & {\bf 49.2} & {\bf 45.1} & {\bf 45.4} & {\bf 27.2} & {\bf 61.4} & {\bf 49.2} & {\bf 33.5}\\
\bottomrule
\end{tabular}
}
\label{tab3}
\end{center}
\end{table*}

\begin{table*}[t]
\begin{center}
\caption{Clustering performance (\%) on fine-grained image clustering datasets. The best results are highlighted in bold.}
\resizebox{\textwidth}{!}{%
\begin{tabular}{l|ccc|ccc|ccc|ccc|ccc}
\toprule
\multicolumn{1}{l|}{\textbf{Dataset}} 
                 & \multicolumn{3}{c|}{\textbf{Aircraft}}   
                 & \multicolumn{3}{c|}{\textbf{Food}} 
                 & \multicolumn{3}{c|}{\textbf{Flowers}}              & \multicolumn{3}{c|}{\textbf{Pets}}          
                 & \multicolumn{3}{c}{\textbf{Cars}}                
                 \\
\midrule
\textbf{Metrics} & \textbf{NMI} & \textbf{ACC} & \textbf{ARI}             
                 & \textbf{NMI} & \textbf{ACC} & \textbf{ARI}             
                 & \textbf{NMI} & \textbf{ACC} & \textbf{ARI}             
                 & \textbf{NMI} & \textbf{ACC} & \textbf{ARI}             
                 & \textbf{NMI} & \textbf{ACC} & \textbf{ARI}             
                 \\
\midrule
TAC (ICML24) & 47.7 & 20.1 & 10.4 & 69.4 & 59.2 & 43.9 & 86.0 & 66.9 & 58.5 & 80.4 & 66.9 & 59.2 & 64.7 & 33.2 & 21.7\\
\rowcolor{LightCyan} GradNorm (ours)  &{\bf 50.3} & {\bf 24.0} & {\bf 13.1} &{\bf 75.0} & {\bf 67.8} & {\bf 52.4} & {\bf 86.7} & {\bf 70.8} & {\bf 64.2} & {\bf 81.5} & {\bf 72.0} & {\bf 62.8} & {\bf 68.3} & {\bf 38.0} & {\bf 26.6}\\
\bottomrule
\end{tabular}
}
\label{tab5}
\end{center}
\end{table*}

% \begin{table*}[t]
% \begin{center}
% \caption{Clustering performance (\%) on five widely used image clustering datasets. The best results are highlighted in bold.}
% \resizebox{0.8\textwidth}{!}{%
% \begin{tabular}{l|ccc|ccc|ccc|ccc}
% \toprule
% \multicolumn{1}{l|}{\textbf{Dataset}} 
%                  & \multicolumn{3}{c|}{\textbf{ImageNet-C~\cite{70}}}   
%                  & \multicolumn{3}{c|}{\textbf{ImageNet-V2~\cite{71}}} 
%                  & \multicolumn{3}{c|}{\textbf{ImageNet-S~\cite{72}}}             
%                  & \multicolumn{3}{c}{\textbf{Average}}               \\
% \midrule
% \textbf{Metrics} & \textbf{NMI} & \textbf{ACC} & \textbf{ARI}             
%                  & \textbf{NMI} & \textbf{ACC} & \textbf{ARI}             
%                  & \textbf{NMI} & \textbf{ACC} & \textbf{ARI}             
%                  & \textbf{NMI} & \textbf{ACC} & \textbf{ARI}                       
%                  \\
% \midrule
% TAC (ICML24) & 71.4 & 39.2 & 25.6 & 71.7 & 38.5 & 23.0 & 70.7 & 34.8 & 22.1 & 71.3 & 37.5 & 23.6 \\
% \rowcolor{LightCyan} GradNorm (ours)  &{\bf 74.2} & {\bf 42.6} & {\bf 27.5} &{\bf 73.6} & {\bf 41.1} & {\bf 26.8} & {\bf 72.4} & {\bf 39.3} & {\bf 24.9} & {\bf 73.4} & {\bf 41.0} & {\bf 26.4}\\
% \bottomrule
% \end{tabular}
% }
% \label{tab6}
% \end{center}
% \end{table*}

\begin{table*}[]
\centering
\caption{Clustering performance (\%) robustness to domain shift. The best results are highlighted in bold.}
\resizebox{0.95\textwidth}{!}{%
\begin{tabular}{l|ccc|ccc|ccc|ccc|ccc} 
\toprule
\textbf{Dataset} & \multicolumn{3}{c|}{\bf ImageNet-C} 
        & \multicolumn{3}{c|}{\bf ImageNet-V2} 
        & \multicolumn{3}{c|}{\bf ImageNet-S} 
        & \multicolumn{3}{c|}{\bf ImageNet-R} 
        & \multicolumn{3}{c}{\bf ImageNet-A}\\
\cmidrule{1-16}
\textbf{Metrics} & \textbf{NMI} & \textbf{ACC} & \textbf{ARI}
                 & \textbf{NMI} & \textbf{ACC} & \textbf{ARI}
                 & \textbf{NMI} & \textbf{ACC} & \textbf{ARI}
                 & \textbf{NMI} & \textbf{ACC} & \textbf{ARI}
                 & \textbf{NMI} & \textbf{ACC} & \textbf{ARI}\\ 
\midrule
TAC (ICML24) & 68.6 & 37.6 & 24.7 
             & 75.9 & 38.0 & 22.6 
             & 17.8 & 32.7 & 18.9 
             & 58.1 & 40.6 & 27.1 
             & 48.0 & 20.6 & 9.9\\
\rowcolor{LightCyan} 
Ours & \textbf{71.9} & \textbf{40.8} & \textbf{26.3} 
     & \textbf{79.6} & \textbf{42.3} & \textbf{25.5} 
     & \textbf{20.8} & \textbf{35.1} & \textbf{22.1} 
     & \textbf{59.2} & \textbf{42.5} & \textbf{28.6} 
     & \textbf{50.6} & \textbf{23.1} & \textbf{11.1}\\
\bottomrule
\end{tabular}%
}
\end{table*}

\subsection{Main results}
\subsubsection{Performance on Classical Datasets}
We evaluate our proposed GradNorm on five widely-used image clustering datasets, compared with 21 deep clustering baselines. Significantly different from early baselines adopt either ResNet-34 or ResNet-18 as the backbone, this paper mainly focuses on comparisons with zero-shot CLIP and CLIP-based methods.
As shown in Table \ref{tab1}, GradNorm consistently outperforms the mostly rececnt TAC~\cite{27} on 5 classic datasets. In particular, GradNorm achieves a notable 7.3\% and 6.1\% improvement in ARI and ACC on ImageNet-Dogs respectively, which eposes its theoretical superiority in Section~\ref{sec4}. While SIC~\cite{26} slightly outperforms GradNorm on the CIFAR-10 dataset, it is worth mentioning that SIC~\cite{26} requires more trainable parameters and a more sophisticated training strategy.
\subsubsection{Performance on Challenging Datasets}
Considering that the rapid development of network pre-training has made clustering on relatively simple datasets such as STL-10 and CIFAR-10 longer challenging, we evaluate GradNorm on three challenging datasets with larger cluster numbers. Table~\ref{tab2} depicts the results on three challenge datasets, where our method still achieves the best performance. To be specific, our GradNorm outperforms TAC~\cite{26} over 5.0\% ACC and 5.2\% ARI on DTD. Besides, our method also outperforms supervised zero-shot CLIP, which highlights the effectiveness of our approach in applying CLIP for clustering tasks. 
\subsection{Visualizations}
To provide an intuitive understanding of our empirical superiority in clustering, we present t-SNE~\cite{64} visualization on various features obtained by our GradNorm. Compared with the pre-trained CLIP image features in Figure~\ref{fig1a} that suffers from remarkable overlapping among image features of different classes, the constructed text counterpart in Figure~\ref{fig1b} exhibit better separation among clusters. Finally, Figure~\ref{fig1b} implies that simply concatenating images and text counterparts could better collaborate the image and text modalities, achieving the best trade-off between within-clustering compactness and between-cluster separation.
\subsection{Ablation Study}
\subsubsection{Ablation on Hyper-parameters}
We evaluate the hyper-parameters most essential to the algorithmic design of our GradNorm.
To assess the impact of the temperature hyper-parameter $\kappa$ in Eq. (\ref{eq9}), we vary the value of $\kappa$ from 0.002 to 0.02. The resulting clustering performance on DTD and CIFAR-20 is reported in Figure~\ref{fig0a} and Figure~\ref{fig0b} respectively. To assess the impact of the temperature hyper-parameter $\tau$ in Eq. (\ref{eq4}) and Eq. (\ref{eq6}), we vary the value of $\tau$ from 5 to 100. The resulting clustering performance on DTD and CIFAR-20 is reported in Figure~\ref{fig1a} and Figure~\ref{fig1b} respectively. As illustrated in Figure~\ref{fig2a} and Figure~\ref{fig2b}, the clustering performance of GradNorm exhibits an initial improvement as $\beta$ increases, followed by either reaching a stable level or degrading slightly when $\beta$ is too high. We suspect that incorporating excessive nouns can introduce unrelated semantics, which has an adverse effect on the clustering process. 
\subsubsection{Ablation on Visual Encoder}
In principle, our GradNorm is generic to the choice of visual encoder. We evaluate GradNorm with different visual encoder architectures, including ViT-B/16 and ViT-L/14, and report the corresponding clustering results in Table~\ref{tab3}. On the one hand, the clustering performance can be enhanced by more powerful visual encoders. On the other hand, GradNorm consistently outperforms TAC regardless of the backbone architecture used, which implies the better generalization of GradNorm over TAC.
\subsection{Extensions}
\subsubsection{Fine-grained Image Clustering}
We validate our method in the fine-grained scenario, where experiments are conducted on five popular datasets including Aircraft~\cite{65}, Food~\cite{66}, Flowers~\cite{67}, Pets~\cite{68}, Cars~\cite{69}. Experiment results on Table~\ref{tab5} shows that our GradNorm consistently outperforms the state-of-the-art TAC, which highlights the superiority of GradNorm in exploring suitable textual semantics for image clustering. 
\subsubsection{Domain-generalizable Image Clustering}
To validate the transferability capability of GradNorm, we perform clustering several versions of ImageNet-1K with diverse domain shifts based on the filtered candidate positive semantics from ImageNet training data.
Experiment results in Table~\ref{tab6} illustrate that the clustering performance of both TAC and GradNorm deteriorates across diverse domain shifts, which indicates the difficulty of image clustering under such conditions. Nevertheless, our proposed GradNorm continues to consistenly outperform TAC cross diverse ID datasets, thus demonstrating its remarkable robustness against diverse domain shifts.
\section{Conclusion}
In this paper, we propose a novel gradient-based framework GradNorm that exploits the unlabeled in-the-wild textual data for LaIC. Theoretically, GradNorm answers the question of how does unlabeled wild data help LaIC by analyzing the separability of truly positive semantics in the wild. Empirically, GradNorm achieves strong performance compared to competitive baselines on various datasets, which echoes our theoretical insights.
Besides, extensive ablations provide further understandings of our GradNorm.
\section*{Acknowledgement}
This work is supported by the Australian Research Council  Australian Laureate Fellowship (FL190100149) and Discovery Early Career Researcher Award (DE250100363).
% {
%     \small
%     \bibliographystyle{ieeenat_fullname}
%     \bibliography{main}
% }
{
\small
\bibliographystyle{ieeenat_fullname}
\bibliography{main,supp}
}

\clearpage
\setcounter{page}{1}
\maketitlesupplementary

\section{Notations and Datasets}
Here we summarize the important notations in Table~\ref{tab7} and the details of datasets in Table~\ref{tab6}.
\begin{table*}[t]
\centering
\caption{Main notations and their descriptions.}
\begin{tabular}{l|c}
\toprule
\textbf{Notation} & \textbf{Description}\\
\midrule
$\Delta$ & Prompt template\\
$f_\mathcal{X}$ & CLIP image encoder\\
$f_\mathcal{T}$ & CLIP text encoder\\
$\mathcal{Z},\mathcal{Y},\mathcal{W}$ & CLIP feature space, Pseudo-label space, Parameter space\\
$h,\mathbf{W}$ & Classifier, Parameters of $h$\\
$\mathcal{D}_\mathcal{X}, N$& Unlabeled image dataset, The size of $\mathcal{D}_\mathcal{X}$\\
$\mathcal{D}_\mathcal{T}, M$ &Unlabeled wild textual dataset, The size of $\mathcal{D}_\mathcal{T}$\\
$\mathcal{P}_\mathcal{T}(k), M_k$ &the ground-truth set of positive semantics whose predicted pseudo-label is $k$, The size of $\mathcal{P}_\mathcal{T}(k)$\\
$\mathbf{x}$ & Unlabeled image\\
$\mathbf{e}$ & CLIP feature of unlabeled image\\
$y$ & Image pseudo-label produced by $k$-means\\
$\tilde{\mathbf{t}}$ & wild textual data\\
$\tilde{\mathbf{r}}$ & CLIP feature of wild textual data\\
$\tilde{y}$ & The predicted pseudo-label of wild textual data from $h$\\
$T_k$ & The filtering threshold for wild text data whose predicted pseudo-label is $k$\\
${\| \cdot \|}_F,{\| \cdot \|}_2$ & Frobenius norm, $L_2$ norm\\
\bottomrule
\end{tabular}
\label{tab7}
\end{table*}
\begin{table*}[t]
\centering
\caption{A summary of datasets used for evaluation.}
\begin{tabular}{l|ccccc}
\toprule
\textbf{Dataset} & \textbf{Training Split} & \textbf{Test Split} & \textbf{\# of Training} & \textbf{\# of Test} & \textbf{\# of Classes} \\
\midrule
STL-10 & Train & Test & 5000 & 8000 & 10 \\
CIFAR-10 & Train & Test & 50000 & 10000 & 10 \\
CIFAR-20 & Train & Test & 50000 & 10000 & 20 \\
ImageNet-10 & Train & Test & 13000 & 500 & 10 \\
ImageNet-Dogs & Train & Test & 19500 & 750 & 15 \\
\midrule
DTD & Train+Val & Test & 3760 & 1880 & 47 \\
UCF-101 & Train & Test & 9537 & 3783 & 101 \\
ImageNet-1K & Train & Test & 1281167 & 50000 & 1000 \\
\bottomrule
\end{tabular}
\label{tab6}
\end{table*}

\section{Derivation of Eq. (6) in Main Content}
\label{app6}
\begin{equation*}
\small
\begin{split}
\left \|\frac{\partial \ell\big(h(\tilde{\mathbf{r}}_i;\mathbf{W}^\star),\tilde{y}_i\big)}{\partial \mathbf{W}^\star} \right\|_F^2
&=\sum_{k=1}^C\left \|\frac{\partial \ell\big(h(\tilde{\mathbf{r}}_i;\mathbf{W}^\star),\tilde{y}_i\big)}{\partial \mathbf{w}_k^\star} \right\|_2^2\\
&=\sum_{k=1}^C\left \|\tau\cdot[\tilde{\pi}_{ik}-\mathbb{I}(k=\tilde{y}_i)]\tilde{\mathbf{r}}_i \right\|_2^2\\
&=\tau^2\cdot\sum_{k=1}^C\left \|(\tilde{\pi}_{ik}-\mathbb{I}(k=\tilde{y}_i)) \right\|^2\\
&=\tau^2\sum_{k\neq y_i}^C\tilde{\pi}_{ij}^2+\tau^2(\max_{j\in[C]}\tilde{\pi}_{ij}-1)^2\\
&=\tau^2\cdot\left(\sum_{k\in[C]}\tilde{\pi}_{ik}^2+1-2\max_{j\in[C]}\tilde{\pi}_{ij}\right),
\end{split}
\end{equation*}
where the last two step holds due to the fact that $\tilde{y}_i=\arg\min_{j\in[C]}\ell\big(h(\tilde{\mathbf{r}}_i;\mathbf{W}^\star),j\big)=\arg\max_{k\in[C]}\tilde{\pi}_{ij}$.

\section{Assumptions, Propositions and Lemmas}
\begin{assumption}[$\gamma$-smoothness] 
\label{A3}
The loss function $\ell(\cdot,\cdot)$ (defined over $\mathcal{Z}\times\mathcal{Y}$) is $\gamma$-smooth such that, for any $\mathbf{z}\in\mathcal{Z}$, $y\in[C]$, and $\mathbf{W}, \mathbf{W}'\in\mathcal{W}$,
\begin{equation}
\nonumber
\left |\ell\big(h(\mathbf{z}_;\mathbf{W}),y\big)-\ell\big(h(\mathbf{z}_;\mathbf{W}'),y\big)\right |\leq\gamma\left \|\mathbf{W}-\mathbf{W}'\right \|_F.
\end{equation}
\end{assumption}
\begin{assumption}[$(\rho,\epsilon,\delta)$-Boundness]
\label{A4}
The parameter space $\mathcal{W}\subset\left\{\mathbf{W}\in\mathbb{R}^{d\times C}:\left\|\mathbf{W}-\mathbf{W}_0\right\|_F\leq \rho\right\}$ is within a Frobenius ball of radius $\rho$ around the given point $\mathbf{W}_0$ that should satisfy the following properties:
\begin{enumerate}
    \item $\sup_{(\mathbf{z},y)\sim\mathbb{P}_{\mathcal{Z}\mathcal{Y}}}\ell\big(h(\mathbf{z};\mathbf{W}_0),y\big)=\epsilon$;
    \item $\sup_{(\mathbf{z},y)\sim\mathbb{P}_{\mathcal{Z}\mathcal{Y}}}\left\|\partial\ell\big(h(\mathbf{z};\mathbf{W}_0),y\big)/\partial\mathbf{W}_0\right\|_F=\delta$.
\end{enumerate}
\end{assumption}
\begin{remark}
It can be easily checked that, for the classifier $h(\cdot;\mathbf{W})$ with softmax output function, the Frobenius norm of the Hessian matrix of the cross-entropy function with regard to the weight matrix $\mathbf{W}$ is bounded given a bounded parameter space. As a results, it is always true that the cross-entropy function is $\gamma$-smooth, therefore justifying the above assumptions.
\end{remark}
\begin{proposition}
\label{P1}
if Assumptions~\ref{A3} and~\ref{A4} holds, we have:
% $$\sup_{\mathbf{W}\in\mathcal{W}}\sup_{(\mathbf{z},y)\sim\mathbb{P}_{\mathcal{Z}\mathcal{Y}}}\left\|\partial\ell\big(h(\mathbf{z};\mathbf{W}),y\big)/\partial\mathbf{W}\right\|_F\leq\sqrt{A_1/2},$$
$$\sup_{\mathbf{W}\in\mathcal{W}}\sup_{(\mathbf{z},y)\sim\mathbb{P}_{{Z}{Y}}}\ell\big(h(\mathbf{z};\mathbf{W}),y\big)\leq A,$$
where $A=\gamma\rho^2+\delta\rho+\epsilon$.
% where $A_1=2(\gamma\rho+\delta)^2$ and $A=\gamma\rho^2+\delta\rho+\epsilon$
\end{proposition}
\begin{proof}
One can prove this by Mean Value Theorem of Integrals easily.
\end{proof}
\begin{proposition}[Self-bounding Property]
\label{P2}
if Assumptions~\ref{A3} and~\ref{A4} holds, for any $\mathbf{W}\in\mathcal{W}$, we have:
\begin{equation}
\left\|\partial\ell\big(h(\mathbf{z};\mathbf{W}),y\big)/\partial\mathbf{W}\right\|_F^2\leq2\gamma\cdot\ell\big(h(\mathbf{z};\mathbf{W}),y\big).
\end{equation}
\end{proposition}
\begin{proof}
The detailed proof of Proposition~\ref{P2} can be found in Appendix B of~\citet{supp_a}.
\end{proof}
\begin{proposition}
\label{P3}
If Assumptions~\ref{A3} and \ref{A4}, for any empirical dataset $\mathcal{D}~\sim\mathbb{P}_{ZY}^{\left|\mathcal{D}\right|}$, we have:
\begin{equation}
\begin{split}
\nonumber
\small
\left\|\mathbb{E}_{(\mathbf{z},y)\in\mathcal{D}}\frac{\partial\ell\big(h(\mathbf{z};\mathbf{W}),y\big)}{\partial\mathbf{W}}\right\|_F^2
\leq2\gamma\mathbb{E}_{(\mathbf{z},y)\in\mathcal{D}}\ell\big(h(\mathbf{z};\mathbf{W}),y\big),
\end{split}
\end{equation}
\begin{equation}
\begin{split}
\nonumber
\small
\left\|\mathbb{E}_{(\mathbf{z},y)\sim\mathbb{P}}\frac{\partial\ell\big(h(\mathbf{z};\mathbf{W}),y\big)}{\partial\mathbf{W}}\right\|_F^2
\leq2\gamma\mathbb{E}_{(\mathbf{z},y)\sim\mathbb{P}}\ell\big(h(\mathbf{z};\mathbf{W}),y\big),
\end{split}
\end{equation}
where we use $\mathbb{P}$ as the abbreviation of $\mathbb{P}_{ZY}$ for brevity.
\end{proposition}
\begin{proof}
Given that the squared Frobenius norm $\left
\|\cdot\right\|_F^2$ is a convex function, Jensen's inequality and Proposition~\ref{P2} imply that
\begin{equation}
\small
\nonumber
\begin{split}
\left\|\mathbb{E}_{(\mathbf{z},y)\in\mathcal{D}}\frac{\partial\ell\big(h(\mathbf{z};\mathbf{W}),y\big)}{\partial\mathbf{W}}\right\|_F^2
&\leq\mathbb{E}_{(\mathbf{z},y)\in\mathcal{D}}\left\|\frac{\partial\ell\big(h(\mathbf{z};\mathbf{W}),y\big)}{\partial\mathbf{W}}\right\|_F^2\\
&\leq\mathbb{E}_{(\mathbf{z},y)\in\mathcal{D}}2\gamma\cdot\ell\big(h(\mathbf{z};\mathbf{W}),y\big)\\
&=2\gamma\cdot\mathbb{E}_{(\mathbf{z},y)\in\mathcal{D}}\ell\big(h(\mathbf{z};\mathbf{W}),y\big)
\end{split}
\end{equation}
\begin{equation}
\small
\nonumber
\begin{split}
\left\|\mathbb{E}_{(\mathbf{z},y)\sim\mathbb{P}}\frac{\partial\ell\big(h(\mathbf{z};\mathbf{W}),y\big)}{\partial\mathbf{W}}\right\|_F^2
&\leq\mathbb{E}_{(\mathbf{z},y)\sim\mathbb{P}}\left\|\frac{\partial\ell\big(h(\mathbf{z};\mathbf{W}),y\big)}{\partial\mathbf{W}}\right\|_F^2\\
&\leq\mathbb{E}_{(\mathbf{z},y)\sim\mathbb{P}}2\gamma\cdot\ell\big(h(\mathbf{z};\mathbf{W}),y\big)\\
&=2\gamma\cdot\mathbb{E}_{(\mathbf{z},y)\sim\mathbb{P}}\ell\big(h(\mathbf{z};\mathbf{W}),y\big).
\end{split}
\end{equation}
\end{proof}
% \begin{lemma}
% \label{L1}
% For any N-sized empirical dataset $\mathcal{D}~\sim\mathbb{P}_{ZY}^{N}$, let us define $\mathbf{W}^\star=\arg\min_{\mathbf{W}\in\mathcal{W}}\mathbb{E}_{(\mathbf{z},y)\in\mathcal{D}}\ell\big(h(\mathbf{z};\mathbf{W}),y\big)$ and $\mathbf{W}^\dagger=\arg\min_{\mathbf{W}\in\mathcal{W}}\mathbb{E}_{(\mathbf{z},y)\sim\mathbb{P}}\ell\big(h(\mathbf{z};\mathbf{W}),y\big)$, with the probability at least $1-\zeta>0$, we have:
% \begin{equation}
% \nonumber
% \begin{split}
% &\mathbb{E}_{(\mathbf{z},y)\in\mathcal{D}}\ell\big(h(\mathbf{z};\mathbf{W}^\star),y\big)\\
% \leq&\mathbb{E}_{(\mathbf{z},y)\sim\mathbb{P}}\ell\big(h(\mathbf{z};\mathbf{W}^\dagger),y\big)+ A\sqrt{\frac{\log(1/\zeta)}{2N}}
% \end{split}
% \end{equation}
% \end{lemma}
\begin{lemma}
\label{L1}
For any empirical dataset $\mathcal{D}~\sim\mathbb{P}^{N}$ and $\mathbf{W}\in\mathcal{W}$, with the probability at least $1-\zeta>0$, we have:
\begin{equation}
\nonumber
\begin{split}
&\mathbb{E}_{(\mathbf{z},y)\in\mathcal{D}}\ell\big(h(\mathbf{z};\mathbf{W}),y\big)\\
\leq&\mathbb{E}_{(\mathbf{z},y)\sim\mathbb{P}}\ell\big(h(\mathbf{z};\mathbf{W}),y\big)+ A\sqrt{\frac{\log(1/\zeta)}{2N}}.
\end{split}
\end{equation}
\end{lemma}
\begin{proof}
Without loss of generality, let
$$
\Omega(\mathbf{W},\mathcal{D})=\mathbb{E}_{(\mathbf{z},y)\in\mathcal{D}}\ell\big(h(\mathbf{z};\mathbf{W}),y\big),
$$
$$
\Omega(\mathbf{W},\mathbb{P})=\mathbb{E}_{(\mathbf{z},y)\sim\mathbb{P}}\ell\big(h(\mathbf{z};\mathbf{W}),y\big).
$$
Given that 
$$
\mathbb{E}_{\mathcal{D}~\sim\mathbb{P}^{N}}\left[\Omega(\mathbf{W},\mathcal{D})\right]=\Omega(\mathbf{W},\mathbb{P}),
$$ 
% we have:
% \begin{equation}
% \nonumber
% \Omega(\mathbf{W}^\star,\mathcal{D})\leq\Omega(\mathbf{W}^\dagger,\mathcal{D}).
% \end{equation}
Hoeffding’s inequality implies that, with the probability at least $1-\zeta>0$, we have: 
\begin{equation}
\nonumber
\begin{split}
\Omega(\mathbf{W}^\star,\mathcal{D})-\Omega(\mathbf{W}^\dagger,\mathbb{P})
&\leq\Omega(\mathbf{W}^\dagger,\mathcal{D})-\Omega(\mathbf{W}^\dagger,\mathbb{P})\\
&\leq A\sqrt{\frac{\log(1/\zeta)}{2N}}.
\end{split}
\end{equation}
\end{proof}
\begin{lemma}
\label{L2}
If Assumptions~\ref{A3} and \ref{A4} holds, for any empirical dataset $\mathcal{D}\sim\mathbb{P}^{N}$ and $\mathbf{W}\in\mathcal{W}$, with the probability at least $1-\zeta>0$, we have:
\begin{equation}
\nonumber
\begin{split}
d_{\mathbf{W}}(\mathcal{D},\mathbb{P})=&\Omega(\mathbf{W},\mathcal{D})-\Omega(\mathbf{W},\mathbb{P})\\
\leq&A\sqrt{\frac{\log(1/\zeta)}{2N}}+U\sqrt{\frac{A(A-\epsilon)D}{N}},
\end{split}
\end{equation}
\begin{equation}
\nonumber
\begin{split}
-d_{\mathbf{W}}(\mathcal{D},\mathbb{P})=&\Omega(\mathbf{W},\mathbb{P})-\Omega(\mathbf{W},\mathcal{D})\\
\leq&A\sqrt{\frac{\log(1/\zeta)}{2N}}+U\sqrt{\frac{A(A-\epsilon)D}{N}},
\end{split}
\end{equation}
where $D$ is the dimension of the parameter space $\mathcal{W}$, $U$ is a uniform constant, and
$$
\Omega(\mathbf{W},\mathcal{D})=\mathbb{E}_{(\mathbf{z},y)\in\mathcal{D}}\ell\big(h(\mathbf{z};\mathbf{W}),y\big),
$$
$$
\Omega(\mathbf{W},\mathbb{P})=\mathbb{E}_{(\mathbf{z},y)\sim\mathbb{P}}\ell\big(h(\mathbf{z};\mathbf{W}),y\big).
$$
\end{lemma}
\begin{proof}
Since it can be easily checked that
\begin{equation}
\nonumber
\mathbb{E}_{D\sim\mathbb{P}^N}\left[d_{\mathbf{W}}(\mathcal{D},\mathbb{P})\right]=0, 
\end{equation}
For any $\mathbf{W}\in\mathcal{W}$ and $\mathbf{W}'\in\mathcal{W}$, Proposition 2.6.1 and Lemma 2.6.8 in \citet{supp_b} imply that
\begin{equation}
\nonumber
\begin{split}
&\left\|d_{\mathbf{W}}(\mathcal{D},\mathbb{P})-d_{\mathbf{W}'}(\mathcal{D},\mathbb{P})\right\|_{\Phi}\\
\leq&\frac{u_0}{\sqrt{N}}\left\|\ell\big(h(\mathbf{z};\mathbf{W}),y\big)-\ell\big(h(\mathbf{z};\mathbf{W}'),y\big)  \right\|_{L^\infty(\mathcal{Z}\times\mathcal{Y})},
\end{split}
\end{equation}
where $||\cdot||_{\Phi}$ is the sub-gaussian norm and $u_0$ is a uniform constant. Therefore, the Dudley’s entropy integral~\cite{supp_b} implies that
\begin{equation}
\nonumber
\begin{split}
&\mathbb{E}_{D\sim\mathbb{P}^N}\sup_{\mathbf{W}\in\mathcal{W}}d_{\mathbf{W}}(\mathcal{D},\mathbb{P})\\
\leq&\frac{u_1}{\sqrt{N}}\int_{0}^{+\infty}\sqrt{\log\Upsilon(\mathcal{F},o,L^\infty)}\mathrm{d}o,
\end{split}
\end{equation}
where $\mathcal{F}=\left\{\ell\big(h(\mathbf{z}|\mathbf{W}),y\big):\mathbf{W}\in\mathcal{W}\right\}$, $u_1$ is anther uniform constant, and $\Upsilon(\mathcal{F},o,||\cdot||_\text{max})$ is the covering number under the $L^\infty$ norm.
Due to the fact that
\begin{equation}
\nonumber
\begin{split}
&\mathbb{E}_{D\sim\mathbb{P}^N}\sup_{\mathbf{W}\in\mathcal{W}}d_{\mathbf{W}}(\mathcal{D},\mathbb{P})\\
\leq&\frac{u_1}{\sqrt{N}}\int_{0}^{+\infty}\sqrt{\log\Upsilon(\mathcal{F},o,L^\infty)}\mathrm{d}o\\
&\frac{u_1}{\sqrt{N}}\int_{0}^{A}\sqrt{\log\Upsilon(\mathcal{F},o,L^\infty)}\mathrm{d}o\\
=&\frac{u_1}{\sqrt{N}}A\int_{0}^{1}\sqrt{\log\Upsilon(\mathcal{F},A\cdot o,L^\infty)}\mathrm{d}o,
\end{split}
\end{equation}
according to the McDiarmid’s Inequality, for any $\mathbf{W}\in\mathcal{W}$, with the probability at least $1-\zeta>0$, we have either
\begin{equation}
\nonumber
\small
\begin{split}
&d_{\mathbf{W}}(\mathcal{D},\mathbb{P})\\
\leq&\frac{u_1}{\sqrt{N}}A\int_{0}^{1}\sqrt{\log\Upsilon(\mathcal{F},A\cdot o,L^\infty)}\mathrm{d}o+A\sqrt{\frac{\log(1/\zeta)}{2N}}    
\end{split}
\end{equation}
or
\begin{equation}
\nonumber
\small
\begin{split}
&-d_{\mathbf{W}}(\mathcal{D},\mathbb{P})\\
\leq&\frac{u_1}{\sqrt{N}}A\int_{0}^{1}\sqrt{\log\Upsilon(\mathcal{F},A\cdot o,L^\infty)}\mathrm{d}o+A\sqrt{\frac{\log(1/\zeta)}{2N}}.    
\end{split}
\end{equation}
Note that $\ell\big(h(\mathbf{z};\mathbf{W}),y\big)$ is $(\gamma\rho+\delta)$-Lipschitz with regard to $\mathbf{W}$ under $||\cdot||_F$. Then 
\begin{equation*}
\begin{split}
&\Upsilon(\mathcal{F},A\cdot o,L^\infty)\\
\leq&\Upsilon(\mathcal{W},A\cdot o/(\gamma\rho+\delta),||\cdot||_\text{F})\\
\leq&(1+\frac{2\rho(\gamma\rho+\delta)}{A\cdot o})^D\\
\leq&(1+\frac{2(A-\epsilon)}{A\cdot o})^D,
\end{split}
\end{equation*}
such that
\begin{equation*}
\begin{split}
&\frac{u_1}{\sqrt{N}}A\int_{0}^{1}\sqrt{\log\Upsilon(\mathcal{F},A\cdot o,L^\infty)}\mathrm{d}o\\
=&\frac{u_1}{\sqrt{N}}A\int_{0}^{1}\sqrt{\log(1+\frac{2(A-\epsilon)}{A\cdot o})^D}\mathrm{d}o\\
=&\frac{u_1}{\sqrt{N}}A\int_{0}^{1}\sqrt{D\log(1+\frac{2(A-\epsilon)}{A\cdot o})}\mathrm{d}o\\
\leq&\frac{u_1}{\sqrt{N}}A\sqrt{D}\int_{0}^{1}\sqrt{\frac{2(A-\epsilon)}{A\cdot o}}\mathrm{d}o\\
=&2\frac{u_1}{\sqrt{N}}A\sqrt{D}\sqrt{\frac{2(A-\epsilon)}{A}}\\
=&U\sqrt{\frac{A(A-\epsilon)D}{N}},
\end{split}
\end{equation*}
where $U=2\sqrt{2}u_1$.
\end{proof}
\begin{lemma}
\label{L3}
If Assumptions~\ref{A3} and \ref{A4} hold, for any empirical dataset $\mathcal{D}\sim\mathbb{P}^{N}$ and $\mathcal{D}'\sim\mathbb{P}^{N'}$, with the probability at least $(1-\zeta)^3>0$, we have
\begin{equation*}
\begin{split}
&\Omega(\mathbf{W}^\star,\mathcal{D}')\\
\leq&\Omega(\mathbf{W}^\dagger,\mathbb{P})+A\sqrt{\frac{\log(1/\zeta)}{2N'}}+U\sqrt{\frac{A(A-\epsilon)D}{N'}}\\
+&2A\sqrt{\frac{\log(1/\zeta)}{2N}}+U\sqrt{\frac{A(A-\epsilon)D}{N}},
\end{split}
\end{equation*}
where $D$ is the dimension of the parameter space $\mathcal{W}$, $U$ is a uniform constant, and
\begin{equation*}
\begin{split}
\mathbf{W}^\star
&=\arg\min_{\mathbf{W}\in\mathcal{W}}\mathbb{E}_{(\mathbf{z},y)\in\mathcal{D}}\ell\big(h(\mathbf{z};\mathbf{W}),y\big)\\
&=\arg\min_{\mathbf{W}\in\mathcal{W}}\Omega(\mathbf{W},\mathcal{D}),
\end{split}
\end{equation*}
\begin{equation*}
\begin{split}
\mathbf{W}^\dagger
&=\arg\min_{\mathbf{W}\in\mathcal{W}}\mathbb{E}_{(\mathbf{z},y)\in\mathbb{P}}\ell\big(h(\mathbf{z};\mathbf{W}),y\big)\\
&=\arg\min_{\mathbf{W}\in\mathcal{W}}\Omega(\mathbf{W},\mathbb{P}),
\end{split}
\end{equation*}
$$
\Omega(\mathbf{W}^\star,\mathcal{D}')=\mathbb{E}_{(\mathbf{z},y)\in\mathcal{D}'}\ell\big(h(\mathbf{z};\mathbf{W}^\star),y\big).
$$
\end{lemma}
\begin{proof}
Given that
\begin{equation*}
\begin{split}
&\Omega(\mathbf{W}^\star,\mathcal{D}')-\Omega(\mathbf{W}^\dagger,\mathbb{P})\\
=&\Omega(\mathbf{W}^\star,\mathcal{D}')-\Omega(\mathbf{W}^\star,\mathbb{P})+\Omega(\mathbf{W}^\star,\mathbb{P})-\Omega(\mathbf{W}^\star,\mathcal{D})\\
+&\Omega(\mathbf{W}^\star,\mathcal{D})-\Omega(\mathbf{W}^\dagger,\mathbb{P})\\
\leq&\Omega(\mathbf{W}^\star,\mathcal{D}')-\Omega(\mathbf{W}^\star,\mathbb{P})+\Omega(\mathbf{W}^\star,\mathbb{P})-\Omega(\mathbf{W}^\star,\mathcal{D})\\
+&\Omega(\mathbf{W}^\dagger,\mathcal{D})-\Omega(\mathbf{W}^\dagger,\mathbb{P})\\
=&d_{\mathbf{W}^\star}(\mathcal{D}',\mathbb{P})-d_{\mathbf{W}^\star}(\mathbb{P},\mathcal{D})+\Omega(\mathbf{W}^\dagger,\mathcal{D})-\Omega(\mathbf{W}^\dagger,\mathbb{P}),
\end{split}
\end{equation*}
Lemmas~\ref{L1} and \ref{L2} imply that, with the probability at least $(1-\zeta)^3>0$, we have all of the following:
\begin{equation}
\nonumber
\begin{split}
d_{\mathbf{W}}(\mathcal{D}',\mathbb{P})
\leq A\sqrt{\frac{\log(1/\zeta)}{2N'}}+U\sqrt{\frac{A(A-\epsilon)D}{N'}},
\end{split}
\end{equation}
\begin{equation}
\nonumber
\begin{split}
-d_{\mathbf{W}^\star}(\mathbb{P},\mathcal{D})
\leq A\sqrt{\frac{\log(1/\zeta)}{2N}}+U\sqrt{\frac{A(A-\epsilon)D}{N}}.
\end{split}
\end{equation}
\begin{equation}
\nonumber
\begin{split}
\Omega(\mathbf{W}^\dagger,\mathcal{D})-\Omega(\mathbf{W}^\dagger,\mathbb{P})
\leq A\sqrt{\frac{\log(1/\zeta)}{2N}}.
\end{split}
\end{equation}
\end{proof}
\begin{lemma}
\label{L4}
If Assumptions~\ref{A3} and \ref{A4} hold, for any empirical dataset $\mathcal{D}\sim\mathbb{P}^{N}$ and $\mathcal{D}'\sim\mathbb{P}^{N'}$, with the probability at least $(1-\zeta)^3>0$, we have
\begin{equation*}
\begin{split}
&\mathbb{E}_{(\mathbf{z},y)\in\mathcal{D}'}\left\|\partial\ell\big(h(\mathbf{z};\mathbf{W}^\star),\hat{y}\big)/\partial\mathbf{W}^\star\right\|_F^2\\
\leq&2\gamma\Omega(\mathbf{W}^\dagger,\mathbb{P})+2\gamma (A\sqrt{\frac{\log(1/\zeta)}{2N'}}+U\sqrt{\frac{A(A-\epsilon)D}{N'}}\\
+&2A\sqrt{\frac{\log(1/\zeta)}{2N}}+U\sqrt{\frac{A(A-\epsilon)D}{N}}),
\end{split}
\end{equation*}
where $D$ is the dimension of the parameter space $\mathcal{W}$, $U$ is a uniform constant, $\hat{y}=\arg\min_{k\in[C]}\ell\big(h(\mathbf{z};\mathbf{W}^\star),k\big)$, and
\begin{equation*}
\begin{split}
\mathbf{W}^\star
&=\arg\min_{\mathbf{W}\in\mathcal{W}}\mathbb{E}_{(\mathbf{z},y)\in\mathcal{D}}\ell\big(h(\mathbf{z};\mathbf{W}),y\big)\\
&=\arg\min_{\mathbf{W}\in\mathcal{W}}\Omega(\mathbf{W},\mathcal{D}).
\end{split}
\end{equation*}
\end{lemma}
\begin{proof}
By Proposition~\ref{P2} and Lemma~\ref{L3}, with the probability at least $(1-\zeta)^3>0$, we have
\begin{equation*}
\begin{split}
&\mathbb{E}_{(\mathbf{z},y)\in\mathcal{D}'}\left\|\partial\ell\big(h(\mathbf{z};\mathbf{W}),\hat{y}\big)/\partial\mathbf{W}\right\|_F^2\\
\leq&\mathbb{E}_{(\mathbf{z},y)\in\mathcal{D}'}2\gamma\cdot\ell\big(h(\mathbf{z};\mathbf{W}),\hat{y}\big)\\
\leq&\mathbb{E}_{(\mathbf{z},y)\in\mathcal{D}'}2\gamma\cdot\ell\big(h(\mathbf{z};\mathbf{W}),y\big)\\
=&2\gamma\Omega(\mathbf{W},\mathcal{D})\\
\leq&2\gamma\Omega(\mathbf{W}^\dagger,\mathbb{P})+2\gamma (A\sqrt{\frac{\log(1/\zeta)}{2N'}}+U\sqrt{\frac{A(A-\epsilon)D}{N'}}\\
+&2A\sqrt{\frac{\log(1/\zeta)}{2N}}+U\sqrt{\frac{A(A-\epsilon)D}{N}}).
\end{split}
\end{equation*}
\end{proof}
\begin{lemma}
\label{L5}
Let us define the ground-truth set of positive semantics from the wild data as $$\mathcal{P}_{\mathcal{T}}(k)=\left\{\tilde{\mathbf{t}}_i\in\mathcal{D}_\mathcal{T}:\tilde{\mathbf{t}}_i\sim\mathbb{P}_\text{pos}~and~k=\arg\max_{j\in[L]}\pi_{ij} \right\}$$ and $\left|\mathcal{P}_{\mathcal{T}}(k)\right|=B_k$. If Assumptions~\ref{A3} and \ref{A4} hold, with the probability at least $(1-\zeta)^3>0$, we have the following:
\begin{equation*}
\begin{split}
&\mathbb{E}_{\tilde{\mathbf{t}}_i\in\mathcal{P}_{\mathcal{T}}(k)}\left\|\partial\ell\big(h(\tilde{\mathbf{r}}_i;\mathbf{W}^\star),\tilde{y}\big)/\partial\mathbf{W}^\star\right\|_F^2\\
\leq&2\gamma\Omega(\mathbf{W}^\dagger,\mathbb{P})+2\gamma (A\sqrt{\frac{\log(1/\zeta)}{2B_k}}+U\sqrt{\frac{A(A-\epsilon)D}{B_k}}\\
+&2A\sqrt{\frac{\log(1/\zeta)}{2N}}+U\sqrt{\frac{A(A-\epsilon)D}{N}}),
\end{split}
\end{equation*}
where $D$ is the dimension of the parameter space $\mathcal{W}$, $U$ is a uniform constant, $\tilde{y}_i=\arg\min_{k\in[C]}\ell\big(h(\tilde{\mathbf{t}}_i;\mathbf{W}^\star),k\big)$, and
\begin{equation*}
\mathbf{W}^\star=\arg\min_{\mathbf{W}\in\mathcal{W}}\frac{1}{N}\sum_{i=1}^N\ell\big(h(\mathbf{e}_i;\mathbf{W}),y_i\big).
\end{equation*}
\end{lemma}
\begin{proof}
Lemma~\ref{L4} directly implies this result.
\end{proof}
\section{Proof of Theorem 1 in Main Content}
\label{app9}
\begin{theorem1} 
Let us define the ground-truth set of positive semantics from the wild data as $$\mathcal{P}_{\mathcal{T}}(k)=\left\{\tilde{\mathbf{t}}_i\in\mathcal{D}_\mathcal{T}:\tilde{\mathbf{t}}_i\sim\mathbb{P}_\text{pos}~and~k=\arg\max_{j\in[L]}\pi_{ij} \right\}$$ and $\left|\mathcal{P}_{\mathcal{T}}(k)\right|=B_k$. If Assumptions~\ref{A3} and \ref{A4} hold, with the probability at least $0.97$, we have the following:
\begin{equation*}
\begin{split}
\text{ERR}_\text{pos}(k)
&\triangleq\frac{\left|\left\{\tilde{\mathbf{t}}_i\in\mathcal{P}_{\mathcal{T}}(k):S(\tilde{\mathbf{t}}_i)>T_k\right\}\right|}{B_k}\\
&\leq\frac{2\gamma}{T_k}\left[\min_{\mathbf{W}\in\mathcal{W}}\Omega(\mathbf{W})+O(\sqrt{\frac{1}{B_k}})+O(\sqrt{\frac{1}{N}})\right],
\end{split}
\end{equation*}
where $O(1/N,1/B_k)\geq0$ is a uniform constant that is positively correlated to $1/N$ and $1/O_k$, and $\Omega(\mathbf{W})=\mathbb{E}_{(\mathbf{z},y)\in\mathbb{P}_{ZY}}\ell\big(h(\mathbf{z};\mathbf{W}),y\big)$ denotes the expected risk.
\end{theorem1}
\begin{proof}
Let $S_k$ be the uniform random variable with $\mathcal{P}_{\mathcal{T}}(k)$ as the support and $S_k(\tilde{\mathbf{t}}_i)=\Phi(\tilde{\mathbf{t}}_i)$ for any $\tilde{\mathbf{t}}_i\in\mathcal{P}_{\mathcal{T}}(k)$, then by the Markov
inequality, we have
\begin{equation*}
\begin{split}
\text{ERR}_\text{pos}(k)
&\triangleq\frac{\left|\left\{\tilde{\mathbf{t}}_i\in\mathcal{P}_{\mathcal{T}}(k):S(\tilde{\mathbf{t}}_i)>T_k\right\}\right|}{B_k}\\
&\leq\frac{1}{T_k}\mathbb{E}_{\tilde{\mathbf{t}}_i\in\mathcal{P}_{\mathcal{T}}(k)}\left[S_k(\tilde{\mathbf{t}}_i)\right].
\end{split}
\end{equation*}
As implied by Lemma~\ref{L5}, with the probability at least $(1-\zeta)^3>0$, we have the following:
\begin{equation*}
\begin{split}
&\mathbb{E}_{\tilde{\mathbf{t}}_i\in\mathcal{P}_{\mathcal{T}}(k)}\left[S_k(\tilde{\mathbf{t}}_i)\right]\\
=&\mathbb{E}_{\tilde{\mathbf{t}}_i\in\mathcal{P}_{\mathcal{T}}(k)}\left\|\partial\ell\big(h(\tilde{\mathbf{r}}_i;\mathbf{W}^\star),\tilde{y}\big)/\partial\mathbf{W}^\star\right\|_F^2\\
\leq&2\gamma\Omega(\mathbf{W}^\dagger,\mathbb{P})+2\gamma (A\sqrt{\frac{\log(1/\zeta)}{2B_k}}+U\sqrt{\frac{A(A-\epsilon)D}{B_k}}\\
+&2A\sqrt{\frac{\log(1/\zeta)}{2N}}+U\sqrt{\frac{A(A-\epsilon)D}{N}}).
\end{split}
\end{equation*}
If we set $\zeta=0.01$, with the probability at least $(1-0.01)^3=0.97$, we have:
\begin{equation*}
\begin{split}
&\frac{\left|\left\{\tilde{\mathbf{t}}_i\in\mathcal{P}_{\mathcal{T}}(k):S(\tilde{\mathbf{t}}_i)>T_k\right\}\right|}{B_k}\\
\leq&\frac{2\gamma}{T_k}\Omega(\mathbf{W}^\dagger,\mathbb{P})+\frac{2\gamma}{T_k}\underbrace{(A\sqrt{\frac{\log10}{B_k}}+U\sqrt{\frac{A(A-\epsilon)D}{B_k}})}_{O(\sqrt{1/B_k})}\\
+&\frac{2\gamma}{T_k}\underbrace{(2A\sqrt{\frac{\log10}{N}}+U\sqrt{\frac{A(A-\epsilon)D}{N}})}_{O(\sqrt{1/N})}.
\end{split}
\end{equation*}
\end{proof}

\end{document}